\documentclass{article}

\usepackage{microtype}
\usepackage{graphicx}
\usepackage{subfigure}
\usepackage{booktabs}
\usepackage{soul}
\usepackage{hyperref}

\usepackage[accepted]{icml2024}

\usepackage{amsmath}
\usepackage{amssymb}
\usepackage{amsthm}
\usepackage[utf8]{inputenc} 
\usepackage[T1]{fontenc}    
\usepackage{hyperref}       
\usepackage{url}            
\usepackage{booktabs}       
\usepackage{amsfonts}       
\usepackage{nicefrac}       
\usepackage{microtype}      
\usepackage{xcolor}         
\usepackage{graphicx}
\usepackage{framed}

\usepackage{caption}
\usepackage{subcaption}
\usepackage{wrapfig}
\usepackage{multirow}
\usepackage{amsfonts}
\usepackage{bm}
\usepackage{verbatim}
\usepackage{longtable}
\usepackage{arydshln}
\usepackage{algorithm}
\usepackage{mathtools, nccmath}

\newcommand{\ours}{\texttt{MobileLLM}}
\newcommand{\ourss}{\texttt{MobileLLM-LS}}

\usepackage[capitalize,noabbrev]{cleveref}

\theoremstyle{plain}

\theoremstyle{definition}

\theoremstyle{remark}

\usepackage[textsize=tiny]{todonotes}

\icmltitlerunning{MobileLLM}

\begin{document}

\twocolumn[
\icmltitle{MobileLLM: Optimizing Sub-billion Parameter Language Models \\ for On-Device Use Cases}

\icmlsetsymbol{equal}{*}

\begin{icmlauthorlist}
\icmlauthor{Zechun Liu}{meta}
\icmlauthor{Changsheng Zhao}{meta}
\icmlauthor{Forrest Iandola}{meta}
\icmlauthor{Chen Lai}{meta}
\icmlauthor{Yuandong Tian}{meta}
\icmlauthor{Igor Fedorov}{meta}
\icmlauthor{Yunyang Xiong}{meta}
\icmlauthor{Ernie Chang}{meta}
\icmlauthor{Yangyang Shi}{meta}
\icmlauthor{Raghuraman Krishnamoorthi}{meta}
\icmlauthor{Liangzhen Lai}{meta}
\icmlauthor{Vikas Chandra}{meta}
\end{icmlauthorlist}

\icmlaffiliation{meta}{Meta}

\icmlcorrespondingauthor{Zechun Liu}{zechunliu@meta.com}
\icmlkeywords{Machine Learning, ICML}

\vskip 0.3in
]
\printAffiliationsAndNotice{} 

\begin{abstract}
This paper addresses the growing need for efficient large language models (LLMs) on mobile devices, driven by increasing cloud costs and latency concerns. We focus on designing top-quality LLMs with fewer than a billion parameters, a practical choice for mobile deployment.
Contrary to prevailing belief emphasizing the pivotal role of data and parameter quantity in determining model quality, our investigation underscores the significance of model architecture for sub-billion scale LLMs. Leveraging deep and thin architectures, coupled with embedding sharing and grouped-query attention mechanisms, we establish a strong baseline network denoted as $\ours{}$, which attains a remarkable 2.7\%/4.3\% accuracy boost over preceding 125M/350M state-of-the-art models. Additionally, we propose an immediate block-wise weight-sharing approach with no increase in model size and only marginal latency overhead. The resultant models, denoted as $\ourss{}$, demonstrate a further accuracy enhancement of 0.7\%/0.8\% than $\ours{}$ 125M/350M. 
Moreover, $\ours{}$ model family shows significant improvements compared to previous sub-billion models on chat benchmarks, and demonstrates close correctness to LLaMA-v2 7B in API calling tasks, highlighting the capability of small models for common on-device use cases. 
\end{abstract}

\section{Introduction}
\label{intro}
\begin{figure}[t!]
    \centering
    \hspace{-2em}\includegraphics[width=1.05\linewidth]{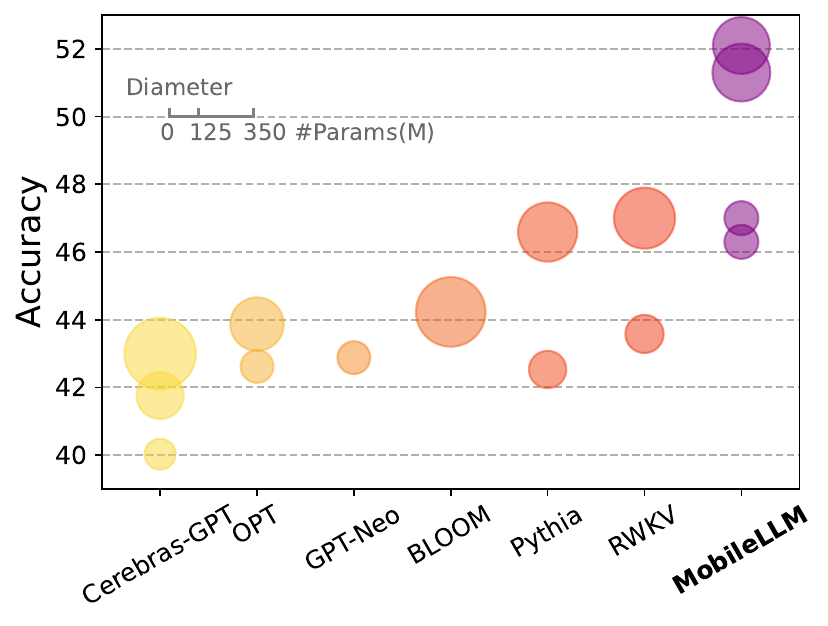}
    \caption{Average score on zero-shot common sense tasks for LLMs smaller than 1B parameters. Each bubble's area is proportional to the model size of a variant in a model family. The results of previous methods were evaluated using open-source Hugging Face models to ensure consistent evaluation procedures. The full list of tasks is in Table~\ref{tab:main}.}
    \label{fig:overall_accuracy}
\end{figure}

Large language models (LLMs) are permeating various facets of human life, influencing not only the way people communicate and work but also shaping everyday entertainment experiences. Prominent examples of contemporary LLM products, such as ChatGPT and Perplexity AI, primarily operate in cloud environments. Leading models such as ChatGPT4 exceed 1 trillion parameters.\footnote{https://the-decoder.com/gpt-4-has-a-trillion-parameters} However, envisioning a future scenario characterized by widespread human reliance on LLMs in both front-end conversational interfaces and back-end operations like recommendation system, equating to $\sim$5\% of individuals' daily time. In this hypothetical scenario, employing GPT-4 at a processing rate of 50 tokens/s entails the deployment of around one hundred million H100 GPUs\footnote{Detailed calculation can be found in the appendix.}, each capable of 60 TFLOPs/s\footnote{https://www.nvidia.com/en-us/data-center/h100/}. This computation scale, excluding communication and data transmission overhead, is on par with 160 Meta-scale companies\footnote{https://twitter.com/soumithchintala/status/1748074223187173724}. The ensuing energy consumption and carbon dioxide emissions would present staggering environmental challenges. Consequently, it is imperative that we downsize LLMs.

\begin{figure}[t!]
    \centering
    \includegraphics[width=\linewidth]{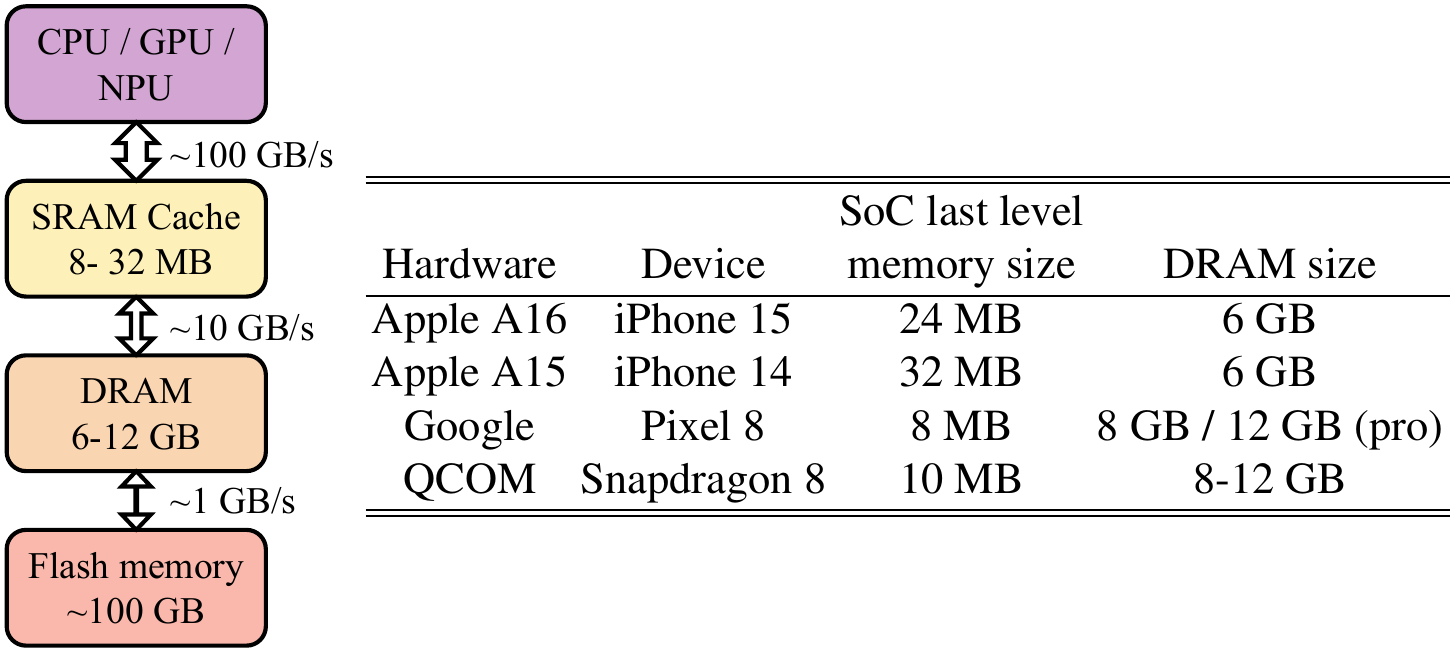}
    \caption{Memory hierarchy in prevalent mobile devices. Despite adequate Flash storage, the operational memory for executing high-speed applications predominantly resides in DRAM, typically constrained to 6-12 GB.}
    \label{fig:hardware}
\end{figure}
Furthermore, considerations of \textit{portability} and \textit{computational cost} propel the necessity to deploy LLMs on smartphones and mobile devices. In the current landscape of mobile technology, integrating an LLM like the LLaMA-v2 7B \cite{touvron2023llama2} with 8-bit weights proves prohibitively expensive due to limitations in main-memory (DRAM) capacity source. A prevalent memory hierarchy in mobile devices is depicted in Figure~\ref{fig:hardware}. 
With DRAM capacities ranging from 6 GB for the iPhone 15 and 12 GB for the Google Pixel 8 Pro \cite{iphone_bionic_A16, google_pixel_8_pro}, a mobile app should not exceed 10\% of the DRAM, since DRAM is shared with the operating system and other applications~\cite{malladi2012towards}. This motivates deploying sub-billion parameter LLMs. Additionally, factoring in LLM energy consumption (0.1 J/token per billion in model parameters~\cite{han2016eie, malladi2012towards}), a 7B-parameter LLM consumes 0.7 J/token. A fully charged iPhone, with approximately 50kJ of energy, can sustain this model in conversation for less than 2 hours at a rate of 10 tokens/s, with every 64 tokens draining 0.2\% of the battery.

These demands converge on a singular imperative: the adoption of compact models for on-device execution. 
By utilizing a sub-billion model, such as a 350M 8-bit model consuming only 0.035 J/token, an iPhone can support conversational use an entire day.
Moreover, the decoding speed can be significantly enhanced, as exemplified by the benchmark results of the 125M model, capable of operating at 50 tokens/s, compared to the state-of-the-art iPhone App MLC Chat utilizing the LLaMA 7B model at 3$\sim$6 tokens/second\footnote{https://llm.mlc.ai}. In light of these considerations, our paper is motivated by the design and implementation of LLMs with parameters less than 1 billion.

We make the following contributions to build the most accurate LLMs to date under 1 billion parameters. \footnote{Our pre-training code is available at \url{https://github.com/facebookresearch/MobileLLM}.}
\begin{itemize}
\vspace{-0.5em}
\item Contradictory to the scaling law~\cite{kaplan2020scaling}, we demonstrate that depth is more important than width for small LLMs. A deep-and-thin model structure excels in capturing abstract concepts, resulting in superior final performance.
\vspace{-0.5em}
\item  We revisit embedding sharing methods~\cite{zhang2022opt} and implement grouped query attention~\cite{ainslie2023grouped_query_attention} in small LLMs to maximize weight utilization.
\vspace{-0.5em}
\item  We propose immediate block-wise weight sharing. In scenarios where memory movement is the latency bottleneck, weight sharing between two adjacent blocks avoids weight movement, requiring only computing the block twice and incurring minimal latency overhead. 
\vspace{-0.5em}
\item We propose a new family of models, $\ours{}$, showcasing SOTA performance. In a suite of zero-shot tasks, $\ours{}$ outperforms the previous SOTA 125M/350M models by 2.7\%/4.3\%. 
\vspace{-0.5em}
\item  In downstream tasks, such as Chat and API calling, $\ours{}$ model family significantly outperforms equivalently-sized models. In an API calling task, $\ours{}$-350M even achieves a comparable exact-match score as a much larger LLaMA-v2 7B model.
\item  We further demonstrate that our design philosophy scales effectively to larger models, with results for $\ours$-600M/1B/1.5B detailed in Appendix~\ref{sec:larger_scale}.
\end{itemize}

\section{Improving Sub-billion Scale LLM Design}
\begin{figure}[t]
    \centering
    \includegraphics[width=\linewidth]{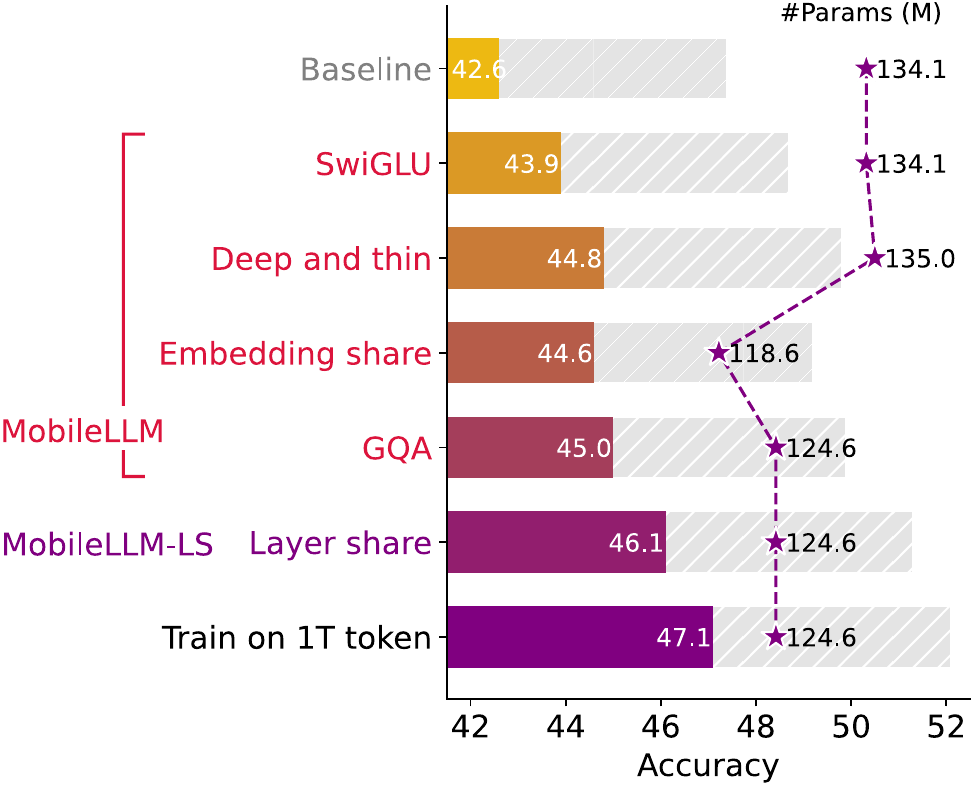}
    \caption{Design roadmap of sub-billion sized transformer models. The foreground and background bars represent the averaged accuracy on zero-shot common sense reasoning tasks for 125M and 350M models, respectively. The 125M model, initially a 12-layer 768-dimension structure, is enhanced via improving feed-forward network design, network depth adjustments, and weight-sharing strategies. The detailed accuracy of each modification can be found in the appendix.} 
    \label{fig:roadmap}
\end{figure}

In this section, we present the evolutionary path from a baseline sub-billion parameter model to the new state-of-the-art models (Figure~\ref{fig:roadmap}). We explore both 125M and 350M models and demonstrate consistent improvements in both cases. For on-device use cases where model size is a major constraint, how to allocate the limited weight parameters effectively becomes more critical than ever. We first propose a strong baseline model named $\ours{}$ by testing out four model design techniques that are beneficial for sub-billion scale LLMs, including (1) adopting SwiGLU FFN ~\cite{dauphin2017language}; (2) forcing \textit{lanky} (deep and thin) architectures (3) revisiting embedding sharing method~\cite{zhang2022opt} (4) utilizing grouped query attention~\cite{chowdhery2023palm}. Then we develop an immediate block-wise layer-sharing method that further boosts accuracy without incurring any additional memory overhead and with only slight latency overhead in the memory-bounded LM decoding process. We denote our model with layer sharing as $\ourss{}$.

\subsection{Training Setup}

Our experiments are conducted on 32 A100 GPUs, with each GPU having a batch size of 32. We performed exploratory experiments with 120k iterations on 0.25T tokens. Subsequently, the top models reported in Table \ref{tab:main} and Table \ref{tab:more_task}, are trained with 480k iterations on 1T tokens.

We evaluate the pre-trained model on zero-shot common sense reasoning tasks, including ARC-easy, ARC-challenge~\cite{clark2018arc}, BoolQ~\cite{clark2019boolq}, PIQA~\cite{bisk2020piqa}, SIQA~\cite{sap2019siqa}, HellaSwag~\cite{zellers2019hellaswag}, OBQA~\cite{mihaylov2018obqa}, WinoGrande~\cite{sakaguchi2021winogrande}, as well as question answering and reading comprehension tasks using TQA~\cite{joshi2017triviaqa} and RACE dataset~\cite{lai2017race}. 

\subsection{Building a Strong Baseline}
\subsubsection{Feed-forward Network Choice}
We first investigate activation functions commonly used in feed-forward networks (FFNs) and find that the state-of-the-art SwiGLU~\cite{dauphin2017language} is also beneficial for small models. By changing the vanilla FFN ($FC \rightarrow ReLU \rightarrow FC$) to SwiGLU, The average performance on zero-shot reasoning tasks is boosted from 42.6 to 43.9 for the 125M model. Therefore, we use SwiGLU in FFN for the experiments afterward.
\subsubsection{Architecture Depth vs Width}
\label{sec:depth_vs_width}
\begin{figure*}[t!]
    \centering
    \includegraphics[width=\linewidth]{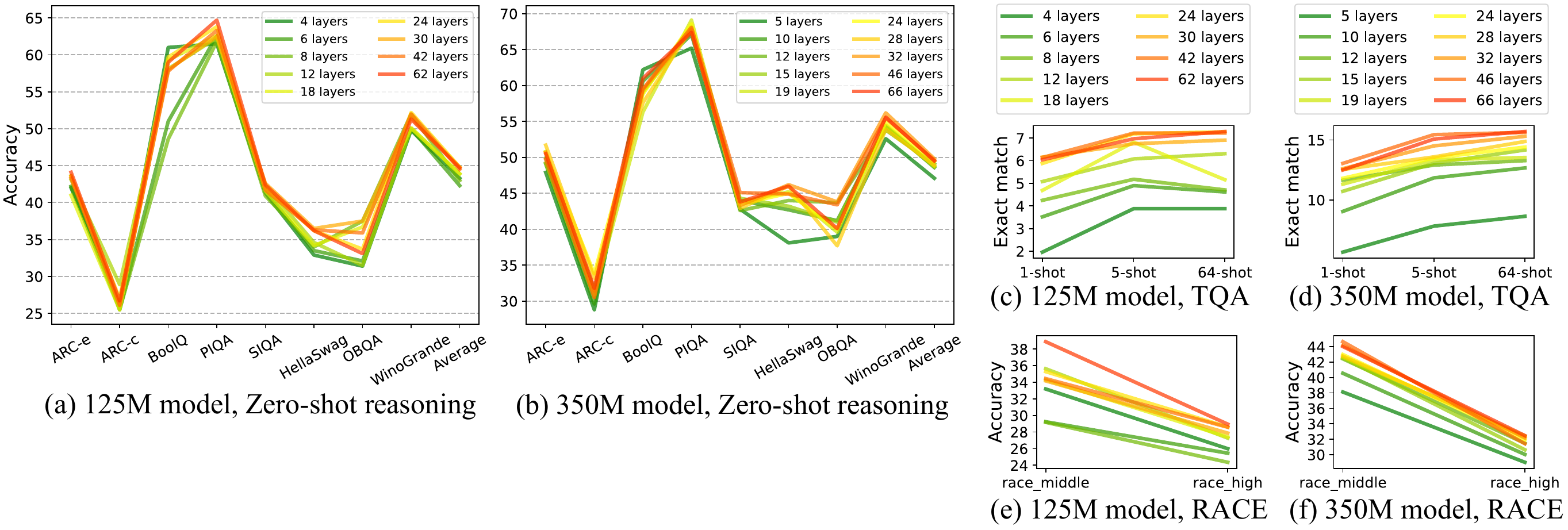}
    \caption{Under comparable model sizes, deeper and thinner models generally outperform their wider and shallower counterparts across various tasks such as zero-shot common sense reasoning, question answering, and reading comprehension.}
    \label{fig:depth_vs_width}
    \vspace{1em}
\end{figure*}
A prevalent belief~\cite{kaplan2020scaling} in the field suggests that the performance of transformer models is primarily determined by the number of parameters, the size of the training dataset, and the number of training iterations. This belief posits that architectural designs have negligible impact on the transformer model's performance. However, our findings indicate that this may not hold true for smaller models.

Our experimental results, specifically for small models with limited model capacity, reveals that going deeper is more crucial than going wider for performance improvement. 
We conducted an extensive study involving the training of 19 models, including 9 models with $\sim$125M parameters and 10 models with $\sim$350M parameters. Each model is designed with a similar size but varied in terms of depth and width. We experiment on eight zero-shot common sense reasoning tasks, as well as question answering and reading comprehension benchmarks. Our findings consistently demonstrate that deeper and thinner models outperform their shallower and wider counterparts. Figure \ref{fig:depth_vs_width} (a) and (b) illustrate the superior performance of deeper networks across most zero-shot reasoning tasks, including ARC-easy, ARC-challenge, PIQA, HellaSwag, OBQA, WinoGrande. Particularly, this trend is even more pronounced on the TQA and RACE datasets, as shown in Figure \ref{fig:depth_vs_width} (c)-(f). Detailed model configurations and results can be seen in the appendix. Our findings suggest that models with 30 or even 42 layers perform significantly better than those with 12 layers for transformer models sized around 125M. This finding is surprising considering the number of layers in most previous 125M models~\cite{zhang2022opt, black2022gptneo} is limited to 12.

\subsubsection{Embedding Sharing}
\begin{table*}[btp]
\renewcommand\arraystretch{0.6}
\centering
\caption{Ablation study on input-output embedding sharing with a 30-layer model with 512 embedding dimension, on zero-shot common-sense reasoning tasks. The increased depth ($\uparrow$ depth) model has 32 layers.}
\label{tab:embeding_share}
\setlength{\tabcolsep}{1.2mm}
{\resizebox{0.85\textwidth}{!}{
\begin{tabular}{lcccccccccccc}
\noalign{\vspace{0.1em}}\hline\noalign{\vspace{0.1em}}
\noalign{\vspace{0.1em}}\hline\noalign{\vspace{0.1em}}
\textbf{Model} & \textbf{\# Params} & \textbf{ARC-e} & \textbf{ARC-c} & \textbf{BoolQ} & \textbf{PIQA} & \textbf{SIQA} & \textbf{HS} & \textbf{OBQA} & \textbf{WinoGrande} & \textbf{Avg.} \\ 
\noalign{\vspace{0.1em}}\hline\noalign{\vspace{0.1em}}
Without emb-share & 135M & 43.6 & 26.1 & 58.0 & 62.5 & 42.6 & 36.5 & 37.5 & 51.5 & 44.8 \\
+ emb-share & 119M & 44.4 & 26.0 & 56.2 & 62.8 & 43.1 & 35.9 & 36.0 & 52.6 & 44.6 \\
+ emb-share, $\uparrow$ depth & 125M & 43.3 & 26.4 & 54.4 & 64.7 & 43.5 & 36.9 & 38.5 & 52.6 & 45.0 \\
\noalign{\vspace{0.1em}}\hline\noalign{\vspace{0.1em}}
\noalign{\vspace{0.1em}}\hline\noalign{\vspace{0.1em}}
\end{tabular}}}
\vspace{1em}
\end{table*}

In sub-billion scale language models, the embedding layers constitute a significant portion of the parameter count. For instance, with an embedding dimension of 512 and a vocabulary size of 32k, the input and output embedding layers each comprise 16 million parameters. Together, these embedding layers account for more than 20\% of the total parameters of a 125M-parameter model.
Contrastingly, this proportion is considerably lower in larger language models. For example, the input and output embeddings only account for  3.7\% of the total number of parameters in the LLaMA-7B model~\cite{touvron2023llama} and a mere 0.7\% in the LLaMA-70B model. This disparity might elucidate why embedding sharing was initially proposed and implemented in OPT models~\cite{zhang2022opt} but was subsequently disregarded in recent designs of LLMs.

In the development of sub-billion scale language models, we revisit the concept of input-output embedding sharing. The input embedding in LLM models maps the token ID in the vocabulary to the corresponding token embedding and has a dimension of $(vocab\_size, embedding\_dim)$. Conversely, the output fully-connected layer maps the embedding dimension back to the logits prediction across the vocabulary, with a weight size of $(vocab\_size, embedding\_dim)$. By sharing the embedding, we reuse the input embedding weights as the output fully connected layer weights, resulting in a more efficient and compact model architecture.

We experiment on a 30-layer 125M model. In Table~\ref{tab:embeding_share}, we demonstrate that sharing the input and output embeddings reduces the number of parameters by 16M, approximately 11.8\% of total parameters with a 0.2 points drop in average accuracy. The marginal accuracy drop can be readily restored by reallocating the saved parameters to add more layers. Increasing the depth to 32 layers produces a 0.4 points accuracy gain while still maintaining 10M fewer parameters compared to the original 135M model. Similar results are also observed in 350M models.
These findings further suggest that embedding sharing is a valuable technique for maximizing weight utilization and optimizing model performance given a limited model storage budget.

\subsubsection{Number of Heads and KV-Heads}
\begin{figure}[t!]
    \hspace{-01em}
    \includegraphics[width=1.05\linewidth]{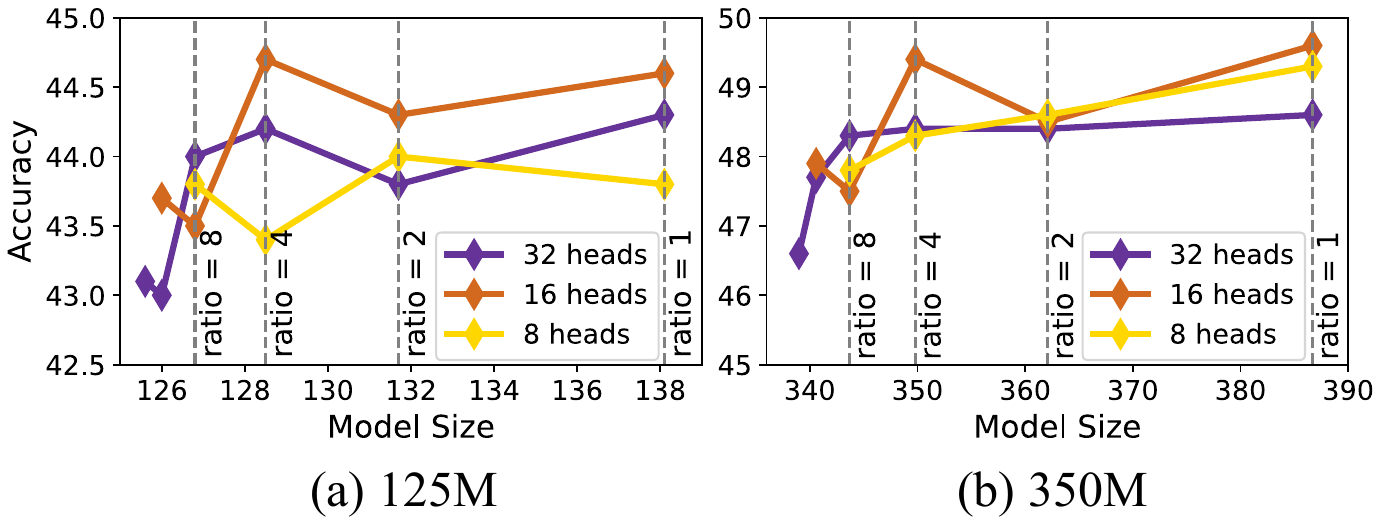}
    \caption{Ablation study on number of heads and kv-heads. Here, the ratio denotes the number of heads divided by the number of kv-heads. Averaged accuracy on zero-shot reasoning tasks is reported. }
    \label{fig:kv_heads}
\end{figure}
We now investigate the optimal head size for small transformer models. The trade-off between more semantics per head dimension and more non-linear combinations of multiple heads is a key consideration in choosing the head size. In addition, most previous studies have typically used an identical number of key-value heads to query heads in sub-billion parameter language models. Instead, we found that grouped query attention, which is initially designed for reducing key-value cache size in LLMs~\cite{chowdhery2023palm, ainslie2023grouped_query_attention}, can also effectively reduce redundancy in key-value heads in small LMs. Grouped query attention can be viewed as another form of weight-sharing for weight re-utilization, where the number of key-value heads is $1/n$ that of query heads, and the kv-heads are repeated $n$ times in computing attention scores and output together with the query. Here, $n \in \mathbb{Z}^{+}$ denotes a positive integer that the number of query heads are divisible by. 

To establish a solid foundation for a state-of-the-art small transformer model, we conducted experiments to determine the desirable head size on 125M and 350M models. Results in Figure \ref{fig:kv_heads} show that using 16 query heads produces the best results. Additionally, reducing the number of kv-heads from 16 to 4 resulted in comparable accuracy for the 125M model and only 0.2 points accuracy drop in the 350M model with almost 10\% model size reduction. These results serve as a guideline in our model architecture design. By adopting the grouped query attention (GQA) and meanwhile increasing the embedding dimension to maintain the model size, the accuracy of 125M further increases by 0.4 points, indicating GQA as a favorable method to further squeeze out small model's potential.

In summary, we tested out four state-of-the-art techniques beneficial to small model designs, including FFN with SwiGLU, deep and thin architecture, embedding sharing, and grouped query attention. Combining these techniques, we build a strong baseline small LLM and we name it $\ours{}$. 

\subsection{Layer Sharing}
\begin{table*}[ht!]
\renewcommand\arraystretch{0.6}
\centering
\caption{Ablation study of layer-sharing strategy on zero-shot common sense reasoning tasks.}
\label{table:layer_share}
\setlength{\tabcolsep}{1.5mm}
{\resizebox{0.95\textwidth}{!}{
\begin{tabular}{llccccccccccc}
\hline\noalign{\vspace{0.1em}}
\noalign{\vspace{0.1em}}\hline\noalign{\vspace{0.1em}}
\textbf{Model} & \textbf{Sharing method} & \textbf{ARC-e} & \textbf{ARC-c} & \textbf{BoolQ} & \textbf{PIQA} & \textbf{SIQA} & \textbf{HellaSwag} & \textbf{OBQA} & \textbf{WinoGrande} & \textbf{Avg.} \\ 
\noalign{\vspace{0.1em}}\hline\noalign{\vspace{0.1em}}
\multirow{4}{*}{125M} & baseline & 41.6 & 25.7 & 61.1 & 62.4 & 43.1 & 34.4 & 36.9 & 51.6 & 44.6 \\ 
& Immediate block-wise share & 43.9 & 27.9 & 61.5 & 64.3 & 41.5 & 35.5 & 35.1 & 50.2 & 45.0 \\ 
& Repeat-all-over share & 43.6 & 27.1 & 60.7 & 63.4 & 42.6 & 35.5 & 36.9 & 51.7 & \textbf{45.2} \\ 
& Reverse share & 43.8 & 26.0 & 58.9 & 62.9 & 42.2 & 35.2 & 36.8 & 52.2 & 44.8 \\ 
\noalign{\vspace{0.1em}}\hdashline[0.8pt/1pt]\noalign{\vspace{0.2em}} 
\multirow{4}{*}{350M} & baseline & 50.8 & 30.6 & 62.3 & 68.6 & 43.5 & 45.1 & 43.8 & 52.4 & 49.6 \\ 
& Immediate block-wise share & 51.5 & 30.8 & 59.6 & 68.2 & 43.9 & 47.7 & 44.7 & 55.0 & 50.2 \\ 
& Repeat-all-over share & 53.5 & 33.0 & 61.2 & 69.4 & 43.2 & 48.3 & 42.2 & 54.6 & \textbf{50.7} \\
& Reverse share & 50.7 & 32.2 & 61.0 & 68.8 & 43.8 & 47.4 & 43.1 & 53.8 & 50.1 \\ 
\noalign{\vspace{0.1em}}\hline\noalign{\vspace{0.1em}}
\noalign{\vspace{0.1em}}\hline\noalign{\vspace{0.1em}}
\end{tabular}}}
\vspace{0.5em}
\end{table*}
The findings in Section \ref{sec:depth_vs_width} on the impact of layer depth versus width suggest deeper layers are favorable for small transformer models. This motivates us to investigate layer sharing as a strategy to increase the number of hidden layers without additional model storage cost. This approach is particularly helpful in on-device scenarios where model size is a major constraint. 

\begin{figure}[t!]
    \centering
    \includegraphics[width=0.8\linewidth]{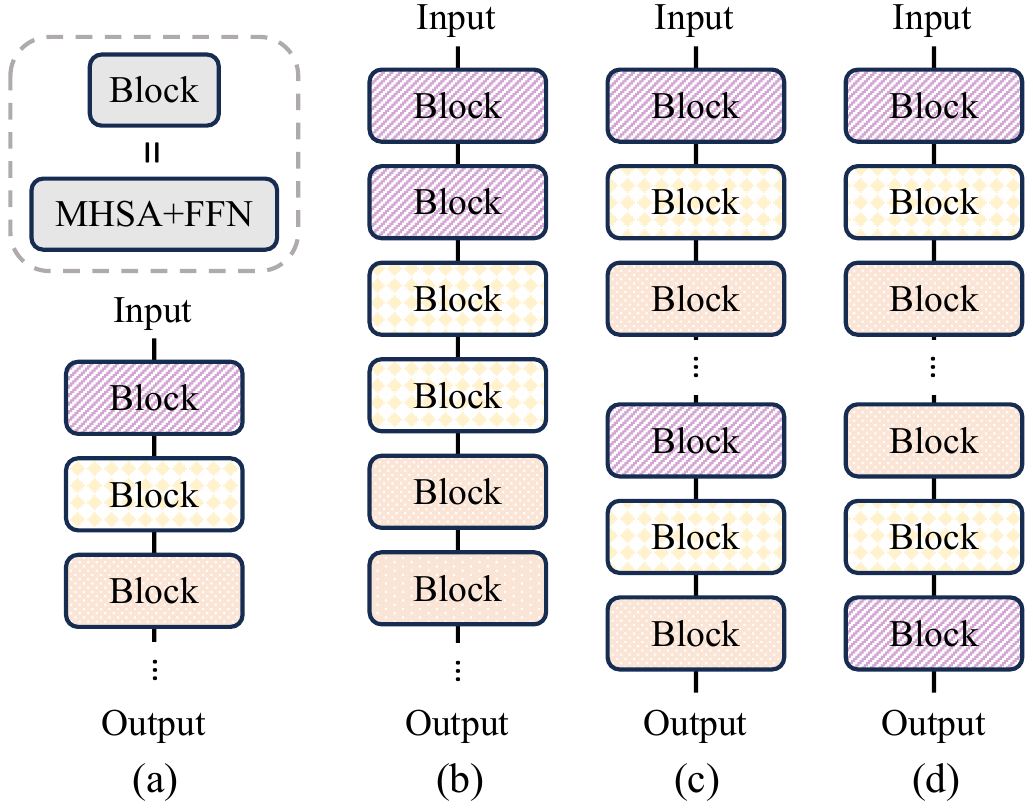}
    \caption{(a) Baseline model without layer sharing; (b) Immediate block-wise sharing; (c) Repeat-all-over sharing; (d) Reverse sharing. A transformer block contains the multi-head self-attention (MHSA) and the feed-forward network (FFN). While repeat-all-over sharing has slightly higher performance, immediate block-wise sharing best utilizes the cache because the shared weights can stay in the cache and be immediately computed twice.}
    \label{fig:layer_share_settings}
\end{figure}

Surprisingly, the experimental findings show that accuracy enhancement can be achieved by simply replicating transformer blocks, without necessitating architectural modifications or an enlargement of the model size.
We further examined three different weight-sharing strategies, illustrated in Figure \ref{fig:layer_share_settings}. Results in Table~\ref{table:layer_share} indicate that the repeat-over layer-sharing strategy produces the best performance among immediate block-wise repeat, repeat all-over, and reverse sharing strategies. However, considering the hardware memory hierarchy (Figure \ref{fig:hardware}), the SRAM for computing is typically limited to around 20MB. This capacity is usually only sufficient to hold a single transformer block. Therefore, placing shared weights in the cache and computing them twice immediately can avoid the need to transfer weights between the SRAM and DRAM, resulting in improved overall execution speed for auto-regressive inference. Consequently, we have opted for the immediate block-wise sharing strategy in our model design. We denote the proposed model with layer sharing as $\ourss{}$.

\section{Experiments}
\subsection{Experimental Settings}
We train $\ours{}$ from scratch using Adam optimizer \cite{kingma2014adam} with a weight decay of 0.1. The experiments are conducted using 32 A100 GPUs, with a batch size of 32 on each GPU. The initial learning rate is set to 2e-3 and follows a cosine learning-rate decay strategy. We perform quick exploration experiments with 120k iterations on 0.25T tokens and train the best models reported in Tables \ref{tab:main} and \ref{tab:more_task} with 480k iterations on 1T tokens. 

\subsection{Main Results}
\begin{table*}[btp]
\renewcommand\arraystretch{0.6}
\centering
\caption{Zero-shot performance on Common Sense Reasoning tasks. $\ours{}$ denotes the proposed baseline model and $\ourss{}$ is integrated with layer sharing with the \#layer counting layers with distinct weights.}
\label{tab:main}
\setlength{\tabcolsep}{1.5mm}
{\resizebox{\textwidth}{!}{
\begin{tabular}{lcccccccccccc}
\noalign{\vspace{0.1em}}\hline\noalign{\vspace{0.1em}}
\noalign{\vspace{0.1em}}\hline\noalign{\vspace{0.1em}}
\textbf{Model} & \textbf{\#Layers} & \textbf{\#Params} & \textbf{ARC-e} & \textbf{ARC-c} & \textbf{BoolQ} & \textbf{PIQA} & \textbf{SIQA} & \textbf{HellaSwag} & \textbf{OBQA} & \textbf{WinoGrande} & \textbf{Avg.} \\ 
\noalign{\vspace{0.1em}}\hline\noalign{\vspace{0.1em}}
Cerebras-GPT-111M & 10 & 111M & 35.8 & 20.2 & \textbf{62.0} & 58.0 & 39.8 & 26.7 & 29.0 & 48.8 & 40.0 \\
LaMini-GPT-124M & 12 & 124M & 43.6 & 26.0 & 51.8 & 62.7 & 42.1 & 30.2 & 29.6 & 49.2 & 41.9 \\
Galactica-125M & 12 & 125M & \textbf{44.0} & 26.2 & 54.9 & 55.4 & 38.9 & 29.6 & 28.2 & 49.6 & 40.9 \\
OPT-125M & 12 & 125M & 41.3 & 25.2 & 57.5 & 62.0 & 41.9 & 31.1 & 31.2 & 50.8 & 42.6 \\
GPT-neo-125M & 12 & 125M & 40.7 & 24.8 & \textbf{61.3} & 62.5 & 41.9 & 29.7 & 31.6 & 50.7 & 42.9 \\
Pythia-160M & 12 & 162M & 40.0 & 25.3 & 59.5 & 62.0 & 41.5 & 29.9 & 31.2 & 50.9 & 42.5 \\
RWKV-169M & 12 & 169M & 42.5 & 25.3 & 59.1 & 63.9 & 40.7 & 31.9 & 33.8 & 51.5 & 43.6 \\
$\ours{}$-125M & 30 & 125M & 43.9 & \textbf{27.1} & 60.2 & \textbf{65.3} & \textbf{42.4} & \textbf{38.9} & \textbf{39.5} & \textbf{53.1} & \textbf{46.3} \\
$\ourss{}$-125M & 30 & 125M & \textbf{45.8} & \textbf{28.7} & 60.4 & \textbf{65.7} & \textbf{42.9} & \textbf{39.5} & \textbf{41.1} & \textbf{52.1} & \textbf{47.0}\\
\noalign{\vspace{0.1em}}\hdashline[0.8pt/1pt]\noalign{\vspace{0.2em}} 
Cerebras-GPT-256M & 14 & 256M & 37.9 & 23.2 & 60.3 & 61.4 & 40.6 & 28.3 & 31.8 & 50.5 & 41.8 \\
OPT-350M & 24 & 331M & 41.9 & 25.7 & 54.0 & 64.8 & 42.6 & 36.2 & 33.3 & 52.4 & 43.9 \\
Pythia-410M & 24 & 405M & 47.1 & 30.3 & 55.3 & 67.2 & 43.1 & 40.1 & 36.2 & 53.4 & 46.6 \\
RWKV-430M & 24 & 430M & 48.9 & 32.0 & 53.4 & 68.1 & 43.6 & 40.6 & 37.8 & 51.6 & 47.0 \\
BLOOM-560M & 24 & 559M & 43.7 & 27.5 & 53.7 & 65.1 & 42.5 & 36.5 & 32.6 & 52.2 & 44.2 \\
Cerebras-GPT-590M & 18 & 590M & 42.6 & 24.9 & 57.7 & 62.8 & 40.9 & 32.0 & 33.2 & 49.7 & 43.0 \\
$\ours{}$-350M & 32 & 345M & \textbf{53.8} & \textbf{33.5} & \textbf{62.4} & \textbf{68.6} & \textbf{44.7} & \textbf{49.6} & \textbf{40.0} & \textbf{57.6} & \textbf{51.3}  \\
$\ourss{}$-350M & 32 & 345M & \textbf{54.4} & \textbf{32.5} & \textbf{62.8} & \textbf{69.8} & \textbf{44.1} & \textbf{50.6} & \textbf{45.8} & \textbf{57.2} & \textbf{52.1}  \\
\noalign{\vspace{0.1em}}\hline\noalign{\vspace{0.1em}}
\noalign{\vspace{0.1em}}\hline\noalign{\vspace{0.1em}}
\end{tabular}}}
\end{table*}
\begin{table}[t]
\renewcommand\arraystretch{0.6}
\centering
\caption{Performance on Trivia QA and RACE datasets for question answering and reading comprehension tasks.}
\label{tab:more_task}
\setlength{\tabcolsep}{1mm}
{\resizebox{0.5\textwidth}{!}{
\begin{tabular}{lcccccccccccc}
\noalign{\vspace{0.1em}}\hline\noalign{\vspace{0.1em}}
\noalign{\vspace{0.1em}}\hline\noalign{\vspace{0.1em}}
 &  \multicolumn{3}{c}{\textbf{TQA} (F1 score)}  &  \multicolumn{2}{c}{\textbf{RACE} (Acc)} \\
\textbf{Model} & 1-shot & 5-shot & 64-shot & middle & high \\
\noalign{\vspace{0.1em}}\hline\noalign{\vspace{0.1em}}
Cerebras-GPT-111M & 1.9 & 3.8 & 4.4 & 29.2 & 24.3 \\
OPT-125M & 8.7 & 9.6 & 8.2 & 34.7 & 27.5 \\
GPT-Neo-125M & 8.0 & 7.9 & 5.0 & 34.7 & 27.0 \\
Pythia-160M & 2.1 & 1.4 & 2.1 & 30.2 & 25.1 \\
$\ours{}$-125M  & \textbf{13.9} & \textbf{14.3} & \textbf{12.5} & \textbf{39.7} & \textbf{28.9} \\
$\ourss{}$-125M & \textbf{14.2} & \textbf{14.8} & \textbf{14.6} & \textbf{40.7} & \textbf{29.6} \\
\noalign{\vspace{0.1em}}\hdashline[0.8pt/1pt]\noalign{\vspace{0.2em}} 
Cerebras-GPT-256M & 5.2 & 6.8 & 3.3 & 31.7 & 26.2 \\
OPT-350M & 11.0 & 12.3 & 10.4 & 37.1 & 28.0 \\
Pythia-410M & 12.4 & 13.8 & 12.8 & 39.1 & 29.7 \\
BLOOM-560M & 8.8 & 8.9 & 5.3 & 37.6 & 28.2 \\
Cerebras-GPT-590M & 6.4 & 9.1 & 4.9 & 34.6 & 27.4 \\
$\ours{}$-350M  & \textbf{22.0} & \textbf{23.9} & \textbf{24.2} & \textbf{45.6} & \textbf{33.8} \\
$\ourss{}$-350M & \textbf{21.4} & \textbf{22.5} & \textbf{22.6} & \textbf{47.3} & \textbf{33.7} \\
\noalign{\vspace{0.1em}}\hline\noalign{\vspace{0.1em}}
\noalign{\vspace{0.1em}}\hline\noalign{\vspace{0.1em}}
\end{tabular}}}
\end{table}

We compare the final performance on zero-shot common sense reasoning tasks, question answering, and reading comprehension tasks. The results of baseline methods were evaluated using their open-source Hugging Face models to ensure consistent evaluation procedures.  

\textbf{Zero-shot Common Sense Reasoning} Table \ref{tab:main} presents a comparison between our proposed model, $\ours{}$, and state-of-the-art sub-billion parameter models, including the early open-sourced LLMs, OPT~\cite{zhang2022opt}, BLOOM~\cite{scao2022bloom}, and recent releases such as Galactica~\cite{taylor2022galactica}, Cerebras~\cite{dey2023cerebras}, GPT-neo~\cite{black2022gptneo} as well as the LLM analyzing suite Pythia~\cite{biderman2023pythia} and transformer variants RWKV~\cite{peng2023rwkv} on zero-shot common sense reasoning tasks.
For the 125M model size, $\ours{}$ favorably outperforms previous models such as OPT, GPT-Neo, and Galactica at the same model size by a significant margin. Additionally, $\ours{}$-125M achieves 3.8 points and 2.7 points higher accuracy than Pythia-160M and RWKV-169M while being 22\% and 26\% smaller, respectively. Furthermore, incorporating layer-sharing in $\ourss{}$-125M results in an additional 0.7 points improvement in accuracy. It is noteworthy that $\ourss{}$-125M achieves comparable or even higher results than most previous 350M models.
In the 350M model size category, $\ours{}$ surpasses previous state-of-the-art models by more than 4 points with comparable or smaller model sizes.

To further validate our design principles under a wider range of memory constraints, we extend our model to include $\ours$-600M, 1B, and 1.5B configurations. The comprehensive results are detailed in the Appendix~\ref{sec:larger_scale}.
 
\textbf{Question Answering and Reading Comprehension} We evaluate pre-trained models on the TQA question answering benchmark~\cite{joshi2017triviaqa} and RACE reading comprehension benchmark~\cite{lai2017race}. We follow the evaluation setup from~\cite{touvron2023llama} and report the results in Table \ref{tab:more_task}. Comparing models of 125M size, $\ours{}$-125M demonstrates a noteworthy improvement of over 4.3 points on the TQA benchmark in contrast to its predecessor. Moreover, the $\ours{}$-350M model exhibits a substantial performance increase of approximately 10 points compared to other 350M-sized models. For the reading comprehension tasks, $\ours{}$ model family also exhibits significantly higher scores than preceding sub-billion parameter models.

\subsection{Downstream Tasks}
To validate the effectiveness of sub-billion scale models for on-device applications, we assess their performance in two crucial on-device tasks: Chat and API calling.
\subsubsection{Chat}
We fine-tune $\ours{}$ models, as well as previous state-of-the-art (SoTA) models sourced from HuggingFace checkpoints for chat-based tasks, and evaluate them under identical settings to ensure consistency. We evaluate two benchmarks: AlpacaEval \cite{alpaca_eval}, a single-run chat benchmark, and MT-Bench \cite{zheng2023mtbench}, a multi-run chat benchmark. The results presented in Table~\ref{tab:chat} showcase that $\ours{}$ models significantly outperform previous state-of-the-art sub-billion scale models, surpassing even models with 1 billion parameters. Notably, $\ourss{}$-350M achieves a remarkable win rate of 48.2\% when compared to the baseline GPT-3 model (text-davinci-001). Considering the self-win rate of GPT-3 is 50\%, it is noteworthy that $\ourss{}$-350M obtains comparable chat performance as this baseline model. Visualizations of chat examples in the appendix also underscore the impressive quality of responses generated by $\ours$ models.

\begin{table}[t]
\centering
\caption{Benchmark results on AlpacaEval (Evaluator: GPT-4; Reference model: text-davinci-001) and MT-Bench.}
\setlength{\tabcolsep}{1mm}
{\resizebox{0.48\textwidth}{!}{
\begin{tabular}{lccccccccccc}
\hline
\noalign{\vspace{0.1em}}\hline\noalign{\vspace{0.1em}}
\textbf{Model} & \begin{tabular}[c]{@{}c@{}}\textbf{MT-Bench}$_\text{(score)}$\end{tabular} & \begin{tabular}[c]{@{}c@{}}\textbf{Alpaca Eval}$_\text{(win \%)}$\end{tabular} \\
\noalign{\vspace{0.1em}}\hline\noalign{\vspace{0.2em}} 
\multicolumn{4}{c}{\textit{number of parameters < 200M}} \\
\noalign{\vspace{0.1em}}\hdashline[0.8pt/1pt]\noalign{\vspace{0.1em}} 
OPT-125M        & 1.21 & 3.91 \\ 
GPT-Neo-125M    & 1.06 & 1.01 \\
Pythia-160M     & 1.01 & 0.63 \\ 
$\ours{}$-125M       & \textbf{2.33} & \textbf{24.07} \\
$\ourss{}$-125M      & \textbf{2.52} & \textbf{23.79} \\ 
\noalign{\vspace{0.1em}}\hline\noalign{\vspace{0.2em}} 
\multicolumn{4}{c}{\textit{200M < number of parameters < 1B}} \\
\noalign{\vspace{0.1em}}\hdashline[0.8pt/1pt]\noalign{\vspace{0.1em}} 
OPT-350M     & 1.37 & 6.80 \\ 
Pythia-410M  & 1.62 & 13.87 \\ 
BLOOM-560M   & 1.73 & 10.29 \\
$\ours{}$-350M    & \textbf{3.28} & \textbf{47.08} \\ 
$\ourss{}$-350M   & \textbf{3.16} & \textbf{48.20} \\
\noalign{\vspace{0.1em}}\hline\noalign{\vspace{0.1em}} 
\multicolumn{4}{c}{\textit{number of parameters > 1B}} \\
\noalign{\vspace{0.1em}}\hdashline[0.8pt/1pt]\noalign{\vspace{0.2em}} 
Pythia-1B    & 1.70 & 16.62 \\
BLOOM-1.1B   & 2.37 & 19.90 \\
Falcon-1.3B  & 2.54 & 30.38\\
OPT-1.3B     & 2.24 & 38.84  \\
\hline
\noalign{\vspace{0.1em}}\hline\noalign{\vspace{0.1em}}
\end{tabular}}}
\label{tab:chat}
\end{table}
\subsubsection{API Calling}

API calling is a common on-device application, particularly in collaboration with audio-to-text models for assistant functionalities. Leveraging LLMs for API calling involves the conversion of natural language inputs into JSON configurations to invoke corresponding APIs\footnote{https://platform.openai.com/docs/guides/function-calling}. For instance, given the input \textit{"Help me set an alarm at 7:30 AM"} the model outputs \textit{\{API: "alarm(time="7:30 am")"\}}. Additionally, the model generates an agent response: \textit{"Sure! Your alarm is set to 7:30 AM."}

To adapt LLMs for this task, we create a synthetic dataset with 5000 training samples and 2500 testing samples. Each sample involves 8 conversation turns on average. Detailed examples of this dataset are provided in the appendix. The pre-trained models undergo fine-tuning on the training set for 4 epochs, utilizing the Adam optimizer with a linear-decay learning rate starting at 2e-5 and a weight decay of 0.01.

Table~\ref{tab:api} shows that $\ours{}$-350M demonstrates comparable intent and structure exact match scores to LLaMA-v2 7B, where high intent scores indicate the correct prediction of the API user intends to call, while structural exact match scores reflect the proficiency in predicting content within API functions. Despite lower Rouge scores in $\ours{}$-350M compared to 7B models, it is crucial to note that API calling prioritizes correct API invocation. The results suggest that certain common scenarios in on-device applications are not particularly challenging, and smaller models like $\ours{}$-350M can adeptly handle it.

\begin{table}[t]
\centering
\caption{API calling evaluation score. EM$_\text{intent}$/EM$_\text{structure}$ measures the exact match in API calling. R1/RL refers to the Rouge-1/-L score measuring the quality of agent response.}
\setlength{\tabcolsep}{1mm}
{\resizebox{0.4\textwidth}{!}{
\begin{tabular}{lccccccccccc}
\hline
\noalign{\vspace{0.1em}}\hline\noalign{\vspace{0.1em}}
\textbf{Model} & \begin{tabular}[c]{@{}c@{}}EM$_\text{intent}$\end{tabular} & \begin{tabular}[c]{@{}c@{}}EM$_\text{structure}$\end{tabular} & R1 & RL \\
\noalign{\vspace{0.1em}}\hdashline[0.8pt/1pt]\noalign{\vspace{0.2em}} 
OPT-350M & 56.1 & 38.6 & 37.1 & 35.3 \\
Pythia-410M & 62.2 & 44.7 & 43.1 & 41.1 \\
BLOOM-560M & \textbf{64.7} & 37.9 & 36.9 & 34.6  \\
$\ours{}$-350M & \textbf{65.3} & \textbf{48.8} & \textbf{46.8} & \textbf{44.6} \\
\noalign{\vspace{0.1em}}\hdashline[0.8pt/1pt]\noalign{\vspace{0.2em}}
LLaMA-v2 7B & 62.8 & \textbf{50.9} & \textbf{56.5} & \textbf{54.3} \\ 
\hline
\noalign{\vspace{0.1em}}\hline\noalign{\vspace{0.1em}}
\end{tabular}}}
\label{tab:api}
\end{table}
\subsection{Compatibility with Quantization}
\label{sec:quantization}
\begin{figure}[t!]
    \centering
    \includegraphics[width=0.8\linewidth]{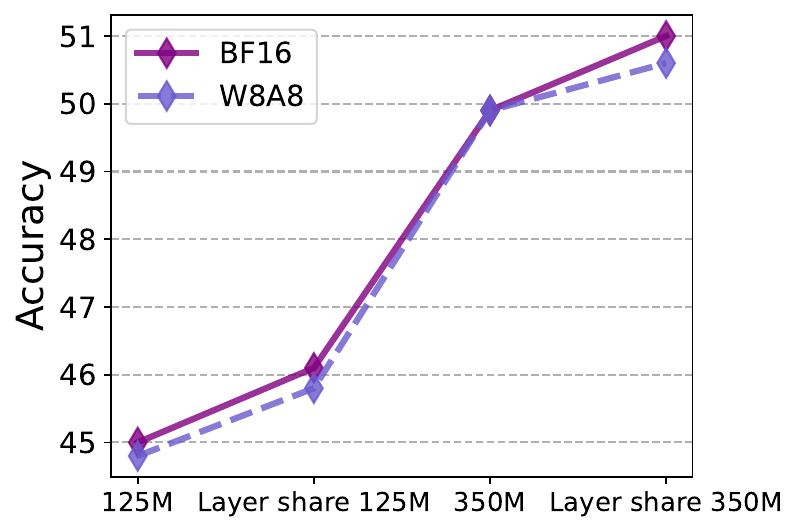}
    \caption{Comparison between BFloat16 model and 8-bit weight 8-bit activation post-training quantized model.}
    \label{fig:quantization}
\end{figure}
We further conduct per-token min-max post-training quantization (PTQ) experiments on both $\ours{}$ and $\ourss{}$ models with 125M and 350M model sizes trained on 0.25T tokens. Figure~\ref{fig:quantization} shows that employing W8A8 PTQ yields a modest accuracy reduction of less than 0.5 points and remains compatible with layer sharing.
\subsection{Knowledge Distillation}
So far, we trained compact models from scratch using next tokens as hard labels.
We explored Knowledge Distillation (KD) of 125M and 350M models with LLAMA-v2 7B as a teacher. 
Unfortunately, KD increases training time (slowdown of $2.6 - 3.2 \times$) and exhibits comparable or inferior accuracy to label-based training (details in appendix). 

\subsection{On-device Profiling}
\begin{table}[t]
\centering
\caption{
Latency analysis of \ours{}-125M (30 layers), \ourss{}-125M (2$\times$30 layers, adjacent blocks sharing weights), and a 60-layer non-shared weight model, with consistent configurations in all other aspects. }
\setlength{\tabcolsep}{2mm}
{\resizebox{0.45\textwidth}{!}{
\begin{tabular}{lccccccccccc}
\hline
\noalign{\vspace{0.1em}}\hline\noalign{\vspace{0.1em}}
& Load & Init & Execute \\
\noalign{\vspace{0.1em}}\hdashline[0.8pt/1pt]\noalign{\vspace{0.2em}}
\ours{} & 39.2 ms & 1361.7 ms & 15.6 ms\\
\ourss{} & 43.6 ms & 1388.2 ms & 16.0 ms \\
60-layer non-shared & 68.6 ms & 3347.7 ms & 29.0 ms\\
\hline
\noalign{\vspace{0.1em}}\hline\noalign{\vspace{0.1em}}
\end{tabular}}}
\label{tab:latency}
\end{table}
We measure the latency for $\ours{}$-125M and $\ourss{}$-125M FP16 models via ExecuTorch\footnote{https://pytorch.org/executorch-overview} on iPhone 13 (iOS 17.2.1), with Metal Performance Shaders (MPS) backend\footnote{https://pytorch.org/executorch/stable/build-run-mps.html}. Model loading, initialization, and execution time are reported in Table~\ref{tab:latency}. Specifically, execution time is averaged over 50 iterations.

Results in Table~\ref{tab:latency} reflects that through weight sharing and doubling the number of layers, $\ourss$ incurs only a 2.2\% increase in loading and initialization time compared to $\ours$, attributable to their similar model sizes. Execution time also experiences a mere 2.6\% overhead, benefitting from data locality. In contrast, a model with a doubled number of layers without weight sharing exhibits a substantial 143\% rise in loading and initialization time and an 86\% increase in execution time.

\section{Related Work}
The excellent performance of LLMs has fostered its wide applications.
Considering the computational cost and energy consumption of LLMs, a new stream of research direction have emerged to downsize LLMs to enable on-device inference. These methods include:

\textbf{Model Compression.} Numerous model compression methods are developed for LLMs, including pruning\cite{xia2023sheared}, sparsity \cite{sun2023simple,xia2023flash,frantar2023sparsegpt}, and quantization \cite{liu2023qllm,dettmers2022llm,kim2023squeezellm, frantar2022gptq, xiao2023smoothquant, yao2022zeroquant, liu2023llm, liu2023llm2, frantar2022gptq}. Our research is complementary to these techniques. As also substantiated in Section~\ref{sec:quantization}, our methodology is compatible with quantization. 

\textbf{Small Model Design.}
A limited number of studies have explored compact model architectures, such as TinyLLaMA \cite{timiryasov2023baby}. However, even the smallest TinyLLaMA exceeds 1 billion parameters, making them still prohibitive for many on-device applications. Some research proposes large model architectures alongside their smaller LLM variants in a model family \cite{zhang2022opt, scao2022bloom, black2022gptneo, dey2023cerebras} or an analytical suite containing small LLM variants~\cite{biderman2023pythia}. However, these models are not optimized under the constraint of sub-billion parameters and therefore may not be optimal. 

\textbf{Neural Architecture Search.} NAS has garnered substantial attention in the realm of convolutional neural networks, particularly in the context of vision tasks \cite{tan2019efficientnet, zoph2016neural, wu2019fbnet, guo2020single}. In contrast, within the transformer domain, the prevailing consensus posits that the model architecture exerts minimal influence on accuracy, provided the total number of parameters remains consistent \cite{kaplan2020scaling}. 
Only a limited number of studies have developed NAS algorithms for language transformers, targeting at BERT models \cite{xu2021bert, jawahar2023mixture, ganesan2021supershaper}. Our current investigation, focusing on the interplay between depth and width, can be conceptualized as a meticulous grid search within the depth space. The outcomes of that study challenge the prevalent orthodoxy surrounding scaling laws, proposing that deep and thin architectures demonstrate higher performance for compact LLMs.

\textbf{Weight Sharing.} Weight sharing is an intuitive strategy for optimizing model weight utilization within fixed parameter constraints. While the OPT family~\cite{zhang2022opt} and subsequent works~\cite{black2022gptneo} leverage weight sharing between input and output embeddings, limited research has explored weight sharing for intermediate layers in transformers~\cite{shen2022sliced,reid2021subformer}. Prior efforts often entail specialized designs for shared layers. In contrast, our contribution highlights a more straightforward yet effective way of simply repeating transformer blocks, yielding improved accuracy with a fixed model size and minimal latency increase.

\textbf{Efficient Attention and Implementation.} In the realm of efficient transformer design, much research has focused on optimizing attention computation through methods like low-rank approximation~\cite{wang2020linformer,katharopoulos2020transformers,xiong2021nystromformer} and sparse attention~\cite{kitaev2020reformer,roy2021efficient}. Another line of work explores hardware scheduling and weight movement, exemplified by works such as FlashAttention~\cite{dao2022flashattention} and FlexGen~\cite{sheng2023flexgen}. In contrast, our primary goal is to optimize model size without introducing new attention computation or efficient hardware implementation methods.

\section{Conclusion}
This study focuses on optimizing sub-billion scale models for on-device applications. Our findings indicate that, for smaller models, prioritizing depth over width enhances model performance. Furthermore, by leveraging advanced weight-sharing techniques, including embedding sharing, grouped query attention, and block-wise weight sharing, we achieve significant enhancements in weight utilization within storage-constrained scenarios. The resulting models denoted as $\ours$ exhibit substantial advancements in zero-shot commonsense reasoning, question answering, and reading comprehension tasks compared to previous SoTA methods. Last but not least, we demonstrate the effectiveness of the fine-tuned $\ours$ models in two prevalent on-device use cases: chat and API calling, underscoring their adeptness in handling such tasks.

\section*{Acknowledgment}
We thank Hansong Zhang for his valuable contribution to setting up the latency measurement environment on iOS, and the full support from the PyTorch edge team.

\section*{Impact Statement}
This paper advocates for the adoption of sub-billion scale large language models in on-device applications, aiming to mitigate energy consumption during LLM inference. The proposed approach is promising in alleviating computational costs associated with LLM deployment.

\bibliography{references}
\bibliographystyle{references}

\newpage
\clearpage
\appendix
\onecolumn
\section*{Appendix}

\section{Scaling Up to Larger Model Architectures}
\label{sec:larger_scale}
In this paper, we primarily investigated two model sizes, $\ours{}$-125M and $\ours{}$-350M. In this section, we extended our design principles—SwiGLU, deeper architecture, grouped-query attention, and embedding sharing—to larger models, pre-training $\ours{}$-600M, 1B, and 1.5B variants. This expansion facilitates a broader range of applications across different memory constraints. Table~\ref{tab:larger_accuracy} compares $\ours{}$ with various general-purpose\footnote{Models pre-trained for specific downstream tasks were excluded to ensure a fair comparison.} pre-trained models, including \textit{i.e.} OPT~\cite{zhang2022opt}, BLOOM~\cite{scao2022bloom}, GPT-neo~\cite{black2022gptneo}, Pythia~\cite{biderman2023pythia}, Falcon~\cite{almazrouei2023falcon}, TinyLlama~\cite{zhang2024tinyllama}, Cerebras-GPT~\cite{dey2023cerebras}, Galactica~\cite{taylor2022galactica}, RWKV~\cite{peng2023rwkv}, LaMini-GPT~\cite{wu2023lamini}, Qwen~\cite{bai2023qwen} as well as a recent model MobiLlama~\cite{thawakar2024mobillama} released after $\ours{}$. 

The results in Table~\ref{tab:larger_accuracy} demonstrate that $\ours{}$ consistently surpasses previous models of similar scale. Notably, $\ours{}$-1.5B achieves an average accuracy of 59.4 points on zero-shot commonsense reasoning tasks, outperforming the previous state-of-the-art model, Qwen1.5-1.8B, by 2.9 points despite the latter having more parameters.
The detailed architecture specifications of $\ours{}$ can be found in Table~\ref{tab:larger_architecture}.

\begin{table*}[h]
\renewcommand\arraystretch{0.6}
\centering
\caption{Zero-shot performance on Common Sense Reasoning tasks for $\ours{}$-600M, 1B and 1.5B. The highest and second-highest average scores within each model-size category are highlighted.}
\label{tab:larger_accuracy}
\setlength{\tabcolsep}{1.5mm}
{\resizebox{0.85\textwidth}{!}{
\begin{tabular}{lcccccccccccc}
\noalign{\vspace{0.1em}}\hline\noalign{\vspace{0.1em}}
\noalign{\vspace{0.1em}}\hline\noalign{\vspace{0.1em}}
\textbf{Model} & \textbf{ARC-e} & \textbf{ARC-c} & \textbf{BoolQ} & \textbf{PIQA} & \textbf{SIQA} & \textbf{HellaSwag} & \textbf{OBQA} & \textbf{WinoGrande} & \textbf{Avg.} \\ 
\noalign{\vspace{0.1em}}\hline\noalign{\vspace{0.1em}}
Qwen1.5-500M & 54.7 & 32.1 & 46.9 & 68.9 & 46.0 &  48.8 & 37.7 & 55.0 & 48.8 \\
BLOOM-560M & 43.7 & 27.5 & 53.7 & 65.1 & 42.5 & 36.5 & 32.6 & 52.2 & 44.2 \\
Cerebras-GPT-590M & 42.6 & 24.9 & 57.7 & 62.8 & 40.9 & 32.0 & 33.2 & 49.7 & 43.0 \\
MobiLlama-800M & 52.0 & 31.7 & 54.6 & 73.0 &  43.3 & 52.3 & 42.5 & 56.3 & \textbf{50.7} \\
$\ours{}$-600M & 58.1 &  35.8 &  61.0 &  72.3 & 44.9 & 55.9 &  47.9 &  58.6 &  \textbf{54.3} \\
\noalign{\vspace{0.1em}}\hdashline[0.8pt/1pt]\noalign{\vspace{0.2em}} 
Pythia-1B & 49.9 & 30.4 & 58.7 & 69.2 & 43.3 & 47.4 & 38.6 & 52.2 & 48.7 \\
MobiLlama-1B & 59.7 & 38.4 & 59.2 & 74.5 & 44.9 & 62.0 & 43.7 & 59.0 & 55.2 \\
Falcon-1B & 59.5 & 38.4 & 63.9 & 74.6 &  44.6 & 62.9 &  45.6 & 60.9 & \textbf{56.3} \\
BLOOM-1.1B & 47.6 & 27.3 & 58.6 & 67.0 & 42.4 & 42.2 & 36.6 & 53.8 & 46.9 \\
TinyLlama-1.1B & 59.2 & 37.1 & 58.1 & 72.9 & 43.9 & 59.1 & 44.7 & 58.8 & 54.2 \\
$\ours{}$-1B & 63.0 &  39.0 &  66.7 &  74.4 & 45.0 &  61.4 & 46.8 &  62.3 &  \textbf{57.3} \\
\noalign{\vspace{0.1em}}\hdashline[0.8pt/1pt]\noalign{\vspace{0.2em}} 
Cerebras-GPT-1.3B & 47.4 & 28.3 & 57.3 & 66.9 & 43.1 & 38.2 & 38.4 & 52.1 & 46.5 \\
Galactica-1.3B & 59.8 & 34.3 & 61.4 & 63.9 & 42.0 & 40.9 & 33.8 & 54.9 & 48.9 \\
GPT-neo-1.3B & 51.3 & 33.0 & 61.8 & 70.9 & 43.7 & 48.6 & 41.2 & 54.5 & 50.6 \\
OPT-1.3B & 54.4 & 31.7 & 58.4 & 71.5 & 44.7 & 53.7 & 44.6 & 59.1 & 52.3 \\
LaMini-GPT-1.5B & 59.9 & 39.1 & 77.0 & 71.9 & 45.9 & 50.9 & 44.4 & 57.5 & 55.8 \\
RWKV-1.5B & 56.2 & 33.8 & 61.8 & 72.3 & 44.7 & 52.8 & 41.8 & 54.7 & 52.3 \\
BLOOM-1.7B & 50.9 & 31.2 & 61.7 & 70.0 & 43.2 & 47.2 & 36.2 & 56.1 & 49.6 \\
Qwen1.5-1.8B  & 61.1 & 36.5 & 68.3 & 74.1 & 47.2 &  60.4 & 42.9 & 61.2 & \textbf{56.5} \\
Cerebras-GPT-2.7B & 53.8 & 32.3 & 55.0 & 71.0 & 43.3 & 48.9 & 40.6 & 55.7 & 50.1 \\
GPT-neo-2.7B & 55.8 & 34.3 & 62.4 & 72.9 & 43.6 & 55.6 & 40.0 & 57.9 & 52.8 \\
OPT-2.7B & 56.6 & 34.6 & 61.8 & 74.5 & 45.6 & 60.2 & 48.2 & 59.6 & 55.1 \\
Pythia-2.8B & 59.4 & 38.9 & 66.1 &  73.8 & 44.5 & 59.6 & 45.0 & 59.4 & 55.8 \\
BLOOM-3B & 55.1 & 33.6 & 62.1 & 70.5 & 43.2 & 53.9 & 41.6 & 58.2 & 52.3 \\
RWKV-3B & 60.1 & 39.1 & 58.6 & 74.5 & 45.1 & 59.8 & 44.6 & 59.1 & 55.1 \\
$\ours{}$-1.5B & 67.5 &  40.9 &  65.7 & 74.8 &  46.4 & 64.5 &  50.5 &  64.7 &  \textbf{59.4} \\
\noalign{\vspace{0.1em}}\hline\noalign{\vspace{0.1em}}
\noalign{\vspace{0.1em}}\hline\noalign{\vspace{0.1em}}
\end{tabular}}}
\end{table*}

\begin{table*}[h]
\renewcommand\arraystretch{0.6}
\centering
\caption{\textbf{Detailed architecture specifications} of MobileLLM. "Emb Dim" denotes the embedding dimension and "Hidden Dim" represents the dimension inside the feed-forward network.}
\label{tab:larger_architecture}
\setlength{\tabcolsep}{1.8mm}
{\resizebox{0.72\textwidth}{!}{
\begin{tabular}{lccccccccccccccccc}
\noalign{\vspace{0.1em}}\hline\noalign{\vspace{0.1em}}
\noalign{\vspace{0.1em}}\hline\noalign{\vspace{0.1em}}
\textbf{Model} & \textbf{\#Layer} & \textbf{\#Head} & \textbf{\#KV-Head} & \textbf{Emb Dim} & \textbf{Hidden Dim} & \textbf{\#Params} \\ 
\noalign{\vspace{0.1em}}\hline\noalign{\vspace{0.1em}} 
$\ours{}$-125M & 30 & 9 & 3 & 576 & 1536 & 124.6M \\
$\ours{}$-350M & 32 & 15 & 5 & 960 & 2560 & 345.3M \\
$\ours{}$-600M & 40 & 18 & 6 & 1152 & 3072 & 603.1M \\
$\ours{}$-1B & 54 & 20 & 5 & 1280 & 3584 & 1.0B \\
$\ours{}$-1.5B & 54 & 25 & 5 & 1600 & 4352 & 1.5B \\
\noalign{\vspace{0.1em}}\hline\noalign{\vspace{0.1em}}
\noalign{\vspace{0.1em}}\hline\noalign{\vspace{0.1em}}
\end{tabular}}}
\end{table*}

\section{Impact of Each Design Choice}

This section presents comprehensive tabulated results for the \textit{improving sub-billion scale LLM design} experiments, at the model sizes of 125M and 350M. Looking at the results in Table~\ref{tab:appendix_roadmap}, transitioning from the traditional Feedforward Network ($FC \rightarrow ReLU \rightarrow FC$) to SwiGLU yields an accuracy improvement of 1.3\% for both model sizes. Further increasing the model depth enhances accuracy by 0.9\%/1.1\% for 125M/350M models, respectively. Then, introducing input and output embedding sharing achieves a parameter reduction of approximately 10\%, while with only marginal accuracy drops of 0.2\% for 125M and 0.6\% for 350M models. Additionally, following in findings in Section~\ref{sec:appendix_kv_head}, we incorporate grouped query attention with a head dimension equal to 64, and a head number to near 4$\times$ to the kv-head number, while increasing the embedding dimension to preserve model size. This modification further results in a performance boost of 0.4\%/0.7\% for 125M/350M models. Combining these techniques establishes a strong baseline network denoted as $\ours{}$. Finally, the immediate block-wise weight-sharing technique contributes an additional accuracy gain of 1.1\% for models trained on 0.25 trillion tokens, resulting in the model $\ourss{}$. Final models including $\ours{}$ and $\ourss{}$ are trained with 1 trillion tokens.

\label{sec:appendix_roadmap}

\begin{table*}[h]
\renewcommand\arraystretch{0.6}
\centering
\caption{Ablation study on the impact of each design choice on the model accuracy on zero-shot common sense reasoning tasks. Corresponding to the bar chart in Figure~\ref{fig:roadmap}. Here, L, H, $\text{H}_\text{KV}$ denotes the number of layers, heads, kv-heads, respectively, and dim denotes the embedding dimension. }
\label{tab:appendix_roadmap}
\setlength{\tabcolsep}{0.7mm}
{\resizebox{0.99\textwidth}{!}{
\begin{tabular}{lccccccccccccccccc}
\noalign{\vspace{0.1em}}\hline\noalign{\vspace{0.1em}}
\noalign{\vspace{0.1em}}\hline\noalign{\vspace{0.1em}}
\textbf{Techniques} & \textbf{L} & \textbf{H} & \textbf{$\text{H}_\text{KV}$} & \textbf{Dim} & \textbf{\#Params(M)} & \textbf{ARC-e} & \textbf{ARC-c} & \textbf{BoolQ} & \textbf{PIQA} & \textbf{SIQA} & \textbf{HellaSwag} & \textbf{OBQA} & \textbf{WinoGrande} & \textbf{Avg.} \\ 
\noalign{\vspace{0.1em}}\hline\noalign{\vspace{0.1em}} 
\textbf{125M} \\
\noalign{\vspace{0.1em}}\hdashline[0.8pt/1pt]\noalign{\vspace{0.2em}} 
Baseline model & 12 & 12 & 12 & 768 & 134.1 & 41.3 & 25.2 & 57.5 & 62.0 & 41.9 & 31.1 & 31.2 & 50.8 & 42.6 \\
+ SwiGLU in FFN & 12 & 12 & 12 & 768 & 134.1 & 43.1 & 28.9 & 58.1 & 62.3 & 42.3 & 34.6 & 31.5 & 50.1 & 43.9 \\
+ Use deep-thin structure & 30 & 8 & 8 & 512 & 135.0 & 43.6 & 26.1 & 58.0 & 62.5 & 42.6 & 36.5 & 37.5 & 51.5 & 44.8 \\
+ Embedding share & 30 & 8 & 8 & 512 & 118.6 & 44.4 & 26.0 & 56.2 & 62.8 & 43.1 & 35.9 & 36.0 & 52.6 & 44.6 \\
+ Grouped-query attention & 30 & 9 & 3 & 576 & 124.6 & 45.5 & 27.7 & 58.3 & 64.6 & 41.9 & 36.4 & 35.4 & 50.4 & 45.0 \\
\ \ \ (Train on 1T token) & 30 & 9 & 3 & 576 & 124.6 & 43.9 & 27.1 & 60.2 & 65.3 & 42.4 & 38.9 & 39.5 & 53.1 & \textbf{46.3} \\
+ Layer sharing & 30 & 9 & 3 & 576 & 124.6 & 44.4 & 27.0 & 61.5 & 65.1 & 43.0 & 37.6 & 37.8 & 52.0 & 46.1 \\
\ \ \ (Train on 1T token) & 30 & 9 & 3 & 576 & 124.6 & 45.8 & 28.7 & 60.4 & 65.7 & 42.9 & 39.5 & 41.1 & 52.1 & \textbf{47.0}\\
\noalign{\vspace{0.1em}}\hline\noalign{\vspace{0.1em}} 
\textbf{350M} \\
\noalign{\vspace{0.1em}}\hdashline[0.8pt/1pt]\noalign{\vspace{0.2em}} 
Baseline model & 15 & 20 & 20 & 1280 & 376.8 & 50.3 & 27.6 & 53.8 & 68.1 & 44.1 & 42.6 & 40.1 & 52.4 & 47.4 \\
+ SwiGLU in FFN & 15 & 20 & 20 & 1280 & 386.7 & 49.2 & 30.6 & 59.1 & 67.7 & 44.3 & 43.2 & 41.0 & 54.2 & 48.7 \\
+ Use deep-thin structure & 32 & 14 & 14 & 896 & 380.3 & 50.7 & 31.4 & 59.4 & 67.8 & 43.3 & 46.2 & 43.8 & 56.2 & 49.8 \\
+ Embedding share & 32 & 14 & 14 & 896 & 351.6 & 49.9 & 32.0 & 60.3 & 67.9 & 43.2 & 47.0 & 38.9 & 54.8 & 49.2 \\
+ Grouped-query attention & 32 & 15 & 5 & 960 & 345.3 & 51.4 & 31.3 & 61.0 & 68.1 & 43.6 & 47.2 & 41.6 & 55.4 & 49.9 \\
\ \ \ (Train on 1T token) & 32 & 15 & 5 & 960 & 345.3 & 53.8 & 33.5 & 62.4 & 68.6 & 44.7 & 49.6 & 40.0 & 57.6 & \textbf{51.3} \\
+ Layer sharing & 32 & 15 & 5 & 960 & 345.3 & 51.9 & 35.2 & 59.6 & 68.9 & 43.4 & 47.2 & 43.3 & 58.4 & 51.0 \\
\ \ \ (Train on 1T token) & 32 & 15 & 5 & 960 & 345.3 & 54.4 & 32.5 & 62.8 & 69.8 & 44.1 & 50.6 & 45.8 & 57.2 & \textbf{52.1} \\
\noalign{\vspace{0.1em}}\hline\noalign{\vspace{0.1em}}
\noalign{\vspace{0.1em}}\hline\noalign{\vspace{0.1em}}
\end{tabular}}}
\end{table*}

\section{Depth vs Width}
\label{sec:appendix_depth_vs_width}

We provide network depth versus width exploration results on zero-shot reasoning tasks in Table~\ref{tab:appendix_depth_vs_width}, as well as results on question answering and reading comprehension tasks in Table~\ref{tab:appendix_depth_more_task}. The findings indicate that shallow architectures with fewer than 10 layers perform poorly in reasoning or handling comprehension tasks. Models with 10-20 layers exhibit improved accuracy, while further increasing depth continues to provide significant benefits across all three tasks. Notably, the optimal depth is found to be near 30 layers for sub-billion scale models.
\begin{table*}[h]
\renewcommand\arraystretch{0.6}
\centering
\caption{Ablation study on depth versus width in architecture design, illustrated in Figure~\ref{fig:depth_vs_width} (a)(b). For compact models, prioritizing depth over width yields superior performance, assessed through zero-shot common sense reasoning tasks.}
\label{tab:appendix_depth_vs_width}
\setlength{\tabcolsep}{1.2mm}
{\resizebox{0.9\textwidth}{!}{
\begin{tabular}{cccccccccccccccc}
\noalign{\vspace{0.1em}}\hline\noalign{\vspace{0.1em}}
\noalign{\vspace{0.1em}}\hline\noalign{\vspace{0.1em}}
\textbf{\#Layer} & \textbf{\#Heads} & \textbf{Dim} & \textbf{\#Params(M)} & \textbf{ARC-e} & \textbf{ARC-c} & \textbf{BoolQ} & \textbf{PIQA} & \textbf{SIQA} & \textbf{HellaSwag} & \textbf{OBQA} & \textbf{WinoGrande} & \textbf{Avg.} \\ 
\noalign{\vspace{0.1em}}\hline\noalign{\vspace{0.1em}}
4 & 20 & 1280 & 163.2 & 42.0 & 26.1 & 61.0 & 61.5 & 41.3 & 32.9 & 31.4 & 49.8 & 43.3 \\
6 & 16 & 1024 & 142.6 & 42.2 & 25.5 & 51.0 & 62.6 & 41.6 & 33.5 & 32.1 & 50.0 & 42.3 \\
8 & 14 & 896 & 138.1 & 43.6 & 25.5 & 48.6 & 61.8 & 40.9 & 34.0 & 37.5 & 52.0 & 43.0 \\
12 & 12 & 768 & 134.1 & 43.1 & 28.9 & 58.1 & 62.3 & 42.3 & 34.6 & 31.5 & 50.1 & 43.9 \\
18 & 10 & 640 & 132.4 & 41.0 & 25.5 & 59.3 & 62.6 & 41.6 & 34.2 & 36.7 & 50.1 & 43.9 \\
24 & 9 & 576 & 132.4 & 43.6 & 27.0 & 59.7 & 63.8 & 42.0 & 36.2 & 33.7 & 52.2 & \textbf{44.8} \\
30 & 8 & 512 & 135.0 & 43.6 & 26.1 & 58.0 & 62.5 & 42.6 & 36.5 & 37.5 & 51.5 & \textbf{44.8} \\
42 & 7 & 448 & 134.6 & 43.3 & 26.1 & 57.8 & 63.3 & 41.7 & 36.3 & 35.9 & 52.0 & 44.5 \\
62 & 6 & 384 & 134.3 & 44.1 & 26.7 & 59.0 & 64.7 & 42.4 & 36.2 & 33.1 & 51.3 & 44.7 \\
\noalign{\vspace{0.1em}}\hdashline[0.8pt/1pt]\noalign{\vspace{0.2em}} 
5 & 32 & 2048 & 388.0 & 47.9 & 28.8 & 62.2 & 65.2 & 42.8 & 38.1 & 39.0 & 52.6 & 47.1 \\
10 & 24 & 1536 & 381.4 & 49.1 & 29.6 & 60.6 & 67.1 & 44.0 & 42.7 & 41.2 & 54.4 & 48.6 \\
12 & 22 & 1408 & 379.9 & 49.9 & 31.6 & 59.1 & 68.2 & 42.6 & 44.0 & 43.7 & 53.8 & 49.1 \\
15 & 20 & 1280 & 386.7 & 49.2 & 30.6 & 59.1 & 67.7 & 44.3 & 43.2 & 41.0 & 54.2 & 48.7 \\
19 & 18 & 1152 & 376.3 & 50.8 & 30.9 & 56.3 & 69.1 & 43.8 & 45.2 & 39.6 & 54.5 & 48.8 \\
24 & 16 & 1024 & 373.8 & 50.7 & 33.8 & 58.8 & 68.6 & 43.5 & 45.0 & 40.0 & 54.9 & 49.4 \\
28 & 15 & 960 & 371.1 & 51.7 & 33.2 & 57.6 & 67.9 & 42.9 & 46.0 & 37.7 & 53.9 & 48.9 \\
32 & 14 & 896 & 380.3 & 50.7 & 31.4 & 59.4 & 67.8 & 43.3 & 46.2 & 43.8 & 56.2 & \textbf{49.8} \\
46 & 12 & 768 & 374.7 & 49.7 & 30.5 & 59.5 & 68.1 & 45.1 & 44.9 & 43.4 & 55.5 & 49.6 \\
66 & 10 & 640 & 376.2 & 50.5 & 31.8 & 61.0 & 67.4 & 43.8 & 46.0 & 40.1 & 55.6 & 49.5 \\
\noalign{\vspace{0.1em}}\hline\noalign{\vspace{0.1em}}
\noalign{\vspace{0.1em}}\hline\noalign{\vspace{0.1em}}
\end{tabular}}}
\end{table*}
\begin{table*}[h!]
\renewcommand\arraystretch{0.6}
\centering
\caption{Ablation study on depth vs. width in architecture design on TQA and RACE datasets, depicted in Figure~\ref{fig:depth_vs_width} (c-f).}
\label{tab:appendix_depth_more_task}
\setlength{\tabcolsep}{1mm}
{\resizebox{0.55\textwidth}{!}{
\begin{tabular}{cccccccccccccccc}
\noalign{\vspace{0.1em}}\hline\noalign{\vspace{0.1em}}
\noalign{\vspace{0.1em}}\hline\noalign{\vspace{0.1em}}
 & & & & \multicolumn{3}{c}{\textbf{TQA} (F1 score)}  &  \multicolumn{2}{c}{\textbf{RACE} (Acc)} \\
\textbf{\#Layer} & \textbf{\#Heads} & \textbf{Dim} & \textbf{\#Params(M)} & 1-shot & 5-shot & 64-shot & middle & high \\
\noalign{\vspace{0.1em}}\hline\noalign{\vspace{0.1em}}
4 & 20 & 1280 & 163.2 & 2.0 & 3.9 & 3.9 & 33.2 & 26.0 \\
6 & 16 & 1024 & 142.6 & 3.5 & 4.9 & 4.6 & 29.2 & 25.5 \\
8 & 14 & 896 & 138.1 & 4.2 & 5.2 & 4.7 & 29.2 & 24.3 \\
12 & 12 & 768 & 134.1  & 5.1 & 6.1 & 6.3 & 35.6 & 27.3 \\
18 & 10 & 640 & 132.4  & 4.7 & 6.8 & 5.2 & 34.3 & 27.4 \\
24 & 9 & 576 & 132.4  & 5.9 & 7.2 & 7.3 & 35.3 & 28.6 \\
30 & 8 & 512 & 135.0  & 6.2 & 6.8 & 6.9 & 34.2 & 27.9 \\
42 & 7 & 448 & 134.6  & 6.1 & 7.2 & 7.2 & 34.4 & 28.6 \\
62 & 6 & 384 & 134.3  & 6.0 & 7.0 & 7.3 & 38.9 & 28.9 \\
\noalign{\vspace{0.1em}}\hdashline[0.8pt/1pt]\noalign{\vspace{0.2em}} 
5  & 32 & 2048 & 388.0  & 5.6 & 7.8 & 8.6 & 38.1 & 29.0  \\
10 & 24 & 1536 & 381.4 & 9.0 & 11.8 & 12.7 & 40.6 & 30.0 \\
12 & 22 & 1408 & 379.9 & 11.7 & 12.9 & 13.3 & 42.4 & 31.5 \\
15 & 20 & 1280 & 386.7 & 10.7 & 13.1 & 14.2 & 42.6 & 30.6 \\
19 & 18 & 1152 & 376.3 & 11.3 & 13.3 & 13.5 & 42.6 & 32.4 \\
24 & 16 & 1024 & 373.8 & 11.8 & 13.5 & 14.4 & 43.0 & 31.9 \\
28 & 15 & 960 & 371.1  & 12.5 & 13.6 & 14.9 & 42.8 & 32.3 \\
32 & 14 & 896 & 380.3  & 12.6 & 14.5 & 15.3 & 44.2 & 32.2 \\
46 & 12 & 768 & 374.7  & 13.0 & 15.4 & 15.6 & 44.7 & 31.4 \\
66 & 10 & 640 & 376.2  & 12.5 & 15.1 & 15.7 & 44.0 & 32.5 \\
\noalign{\vspace{0.1em}}\hline\noalign{\vspace{0.1em}}
\noalign{\vspace{0.1em}}\hline\noalign{\vspace{0.1em}}
\end{tabular}}}
\end{table*}

\section{Number of Heads and Key-Value Heads}
\label{sec:appendix_kv_head}
We provide detailed experimental results assessing the impact of the number of attention heads and key-value heads on zero-shot reasoning accuracy in Table~\ref{tab:appendix_kv_heads}. Our study involves two baseline architectures: an 8-layer 125M model with an embedding dimension of 896, and a 15-layer 350M model with an embedding dimension of 1280. We conduct head size sweeps in \{8, 16, 32\}. The findings shown in Table~\ref{tab:appendix_kv_heads} indicate that using 16 heads, with a head dimension close to 64, and 4 key-value heads, yields the best accuracy and memory trade-off. This setting serves as a guiding principle in our model architecture design.
\begin{table*}[!h]
\renewcommand\arraystretch{0.6}
\centering
\caption{Ablation study investigating the impact of the number of attention heads and key-value heads.}
\label{tab:appendix_kv_heads}
\setlength{\tabcolsep}{1mm}
{\resizebox{0.99\textwidth}{!}{
\begin{tabular}{cccccccccccccc}
\noalign{\vspace{0.1em}}\hline\noalign{\vspace{0.1em}}
\noalign{\vspace{0.1em}}\hline\noalign{\vspace{0.1em}}
\textbf{Model} & \textbf{\#Heads} & \textbf{\#KV-Heads} & \textbf{\#Params(M)} & \textbf{ARC-e} & \textbf{ARC-c} & \textbf{BoolQ} & \textbf{PIQA} & \textbf{SIQA} & \textbf{HellaSwag} & \textbf{OBQA} & \textbf{WinoGrande} & \textbf{Avg.} \\ 
\noalign{\vspace{0.1em}}\hline\noalign{\vspace{0.1em}}
& 32 & 32 & 138.1 & 42.1 & 26.9 & 58.4 & 62.2 & 42.1 & 33.8 & 36.7 & 52.4 & 44.3 \\
& 32 & 16 & 131.7 & 41.2 & 26.4 & 57.7 & 62.5 & 42.3 & 33.3 & 34.2 & 52.9 & 43.8 \\
& 32 & 8 & 128.5 & 42.6 & 27.3 & 61.1 & 61.9 & 41.9 & 32.2 & 35.0 & 52.0 & 44.2 \\
& 32 & 4 & 126.8 & 43.1 & 26.8 & 59.8 & 62.7 & 41.4 & 32.5 & 34.4 & 51.1 & 44.0 \\
& 32 & 2 & 126.0 & 39.8 & 26.7 & 59.4 & 59.4 & 42.0 & 31.3 & 32.6 & 52.9 & 43.0 \\
& 32 & 1 & 125.6 & 41.0 & 24.3 & 59.1 & 60.8 & 41.2 & 31.4 & 35.4 & 52.0 & 43.1 \\
\textbf{125M} & 16 & 16 & 138.1 & 41.6 & 25.7 & 61.1 & 62.4 & 43.1 & 34.4 & 36.9 & 51.6 & \textbf{44.6} \\
\# layers=8   & 16 & 8 & 131.7 & 42.4 & 26.4 & 60.7 & 63.4 & 41.9 & 33.5 & 34.7 & 51.5 & 44.3 \\
dim=896   & 16 & 4 & 128.5 & 42.5 & 25.6 & 62.3 & 62.4 & 41.8 & 33.0 & 35.9 & 54.5 & \textbf{44.7} \\
& 16 & 2 & 126.8 & 41.7 & 25.3 & 56.9 & 61.7 & 42.0 & 32.9 & 32.6 & 54.5 & 43.5 \\
& 16 & 1 & 126.0 & 40.4 & 26.3 & 61.8 & 63.2 & 41.7 & 32.0 & 34.0 & 50.4 & 43.7 \\
& 8 & 8 & 138.1 & 41.4 & 25.0 & 58.3 & 61.7 & 41.7 & 33.3 & 35.9 & 53.2 & 43.8 \\
& 8 & 4 & 131.7 & 43.3 & 28.2 & 58.3 & 61.8 & 42.8 & 33.8 & 30.9 & 53.0 & 44.0 \\
& 8 & 2 & 128.5 & 40.7 & 26.2 & 58.1 & 61.2 & 41.6 & 32.8 & 34.8 & 51.5 & 43.4 \\
& 8 & 1 & 126.8 & 42.5 & 24.8 & 59.4 & 62.3 & 42.0 & 32.0 & 36.3 & 51.3 & 43.8 \\
\noalign{\vspace{0.1em}}\hdashline[0.8pt/1pt]\noalign{\vspace{0.2em}} 
& 32 & 32 & 386.7 & 48.6 & 30.4 & 59.7 & 67.2 & 43.9 & 44.0 & 40.9 & 53.9 & 48.6 \\
& 32 & 16 & 362.1 & 48.9 & 31.6 & 57.6 & 68.4 & 43.4 & 43.8 & 38.6 & 54.9 & 48.4 \\
& 32 & 8 & 349.8 & 48.3 & 33.1 & 61.0 & 67.2 & 42.6 & 42.1 & 39.0 & 53.9 & 48.4 \\
& 32 & 4 & 343.7 & 47.2 & 29.8 & 59.4 & 67.2 & 43.5 & 42.5 & 42.5 & 54.1 & 48.3 \\
& 32 & 2 & 340.6 & 47.6 & 30.3 & 62.4 & 66.9 & 42.6 & 41.6 & 38.6 & 52.0 & 47.7 \\
& 32 & 1 & 339.0 & 48.5 & 27.3 & 56.3 & 67.1 & 42.9 & 40.9 & 36.7 & 53.3 & 46.6 \\
\textbf{350M }& 16 & 16 & 386.7 & 50.8 & 30.6 & 62.3 & 68.6 & 43.5 & 45.1 & 43.8 & 52.4 & \textbf{49.6} \\
\# layers=15  & 16 & 8 & 362.1 & 48.5 & 30.7 & 59.4 & 67.3 & 43.8 & 43.8 & 41.3 & 53.3 & 48.5 \\
dim=1280  & 16 & 4 & 349.8 & 49.9 & 30.6 & 60.0 & 69.2 & 43.5 & 44.2 & 41.8 & 55.8 & \textbf{49.4} \\
& 16 & 2 & 343.7 & 49.3 & 28.4 & 55.0 & 67.3 & 42.7 & 42.6 & 40.3 & 54.5 & 47.5 \\
& 16 & 1 & 340.6 & 49.2 & 29.3 & 58.8 & 67.4 & 43.5 & 42.1 & 39.9 & 52.8 & 47.9 \\
& 8 & 8 & 386.7 & 51.0 & 33.2 & 58.2 & 67.0 & 43.9 & 43.9 & 42.6 & 54.7 & 49.3 \\
& 8 & 4 & 362.1 & 50.0 & 31.3 & 60.2 & 66.0 & 42.7 & 43.7 & 40.9 & 53.7 & 48.6 \\
& 8 & 2 & 349.8 & 49.3 & 30.4 & 59.9 & 67.9 & 42.8 & 43.5 & 38.9 & 53.8 & 48.3 \\
& 8 & 1 & 343.7 & 48.0 & 27.7 & 61.5 & 67.1 & 43.1 & 42.5 & 37.3 & 54.7 & 47.8 \\
\noalign{\vspace{0.1em}}\hline\noalign{\vspace{0.1em}}
\noalign{\vspace{0.1em}}\hline\noalign{\vspace{0.1em}}
\end{tabular}}}
\end{table*} 

\section{Layer-Sharing Number Ablation}
\begin{table*}[h!]
\renewcommand\arraystretch{0.6}
\centering
\caption{Ablation study on the impact of varying layer repetition numbers.}
\label{tab:appendix_layer_repeat}
\setlength{\tabcolsep}{1.5mm}
{\resizebox{0.99\textwidth}{!}{
\begin{tabular}{lccccccccccccc}
\noalign{\vspace{0.1em}}\hline\noalign{\vspace{0.1em}}
\noalign{\vspace{0.1em}}\hline\noalign{\vspace{0.1em}}
\textbf{Model Size} & \textbf{Repeat Times} & \textbf{ARC-e} & \textbf{ARC-c} & \textbf{BoolQ} & \textbf{PIQA} & \textbf{SIQA} & \textbf{HellaSwag} & \textbf{OBQA} & \textbf{WinoGrande} & \textbf{Avg.} \\
\noalign{\vspace{0.1em}}\hline\noalign{\vspace{0.1em}}
\multirow{4}{*}{125M} & baseline & 41.6 & 25.7 & 61.1 & 62.4 & 43.1 & 34.4 & 36.9 & 51.6 & 44.6 \\
& block-wise share repeat $\times$2 & 43.9 & 27.9 & 61.5 & 64.3 & 41.5 & 35.5 & 35.1 & 50.2 & 45.0 \\
& block-wise share repeat $\times$3 & 43.3 & 27.6 & 60.9 & 63.0 & 42.3 & 35.6 & 34.1 & 52.9 & 45.0 \\
& block-wise share repeat $\times$4 & 42.0 & 26.6 & 62.3 & 64.4 & 41.5 & 36.2 & 35.8 & 53.9 & 45.3 \\
\noalign{\vspace{0.1em}}\hdashline[0.8pt/1pt]\noalign{\vspace{0.2em}} 
\multirow{4}{*}{350M} & baseline & 50.8 & 30.6 & 62.3 & 68.6 & 43.5 & 45.1 & 43.8 & 52.4 & 49.6 \\
& block-wise share repeat $\times$2 & 51.5 & 30.8 & 59.6 & 68.2 & 43.9 & 47.7 & 44.7 & 55.0 & 50.2 \\
& block-wise share repeat $\times$3 & 49.6 & 32.0 & 57.3 & 68.4 & 43.9 & 47.0 & 40.6 & 56.2 & 49.4 \\
& block-wise share repeat $\times$4 & 52.2 & 32.6 & 60.0 & 69.0 & 44.5 & 47.6 & 41.8 & 55.2 & 50.4 \\
\noalign{\vspace{0.1em}}\hline\noalign{\vspace{0.1em}}
\noalign{\vspace{0.1em}}\hline\noalign{\vspace{0.1em}}
\end{tabular}}}
\end{table*}
We extended our investigation to determine the optimal number of layer repetitions. The experiment involved the 8-layer 125M model with an embedding dimension of 896 and the 15-layer 350M model with an embedding dimension of 1280. The results in Table~\ref{tab:appendix_layer_repeat} demonstrate that when we double the layer number and with each two transformer blocks sharing weights, the accuracy is enhanced by 0.4-0.6\%. However, as we further triple or quadruple the layer repetition times, this accuracy enhancement effect diminishes. Consequently, in our experiments, we adopt a configuration where every two blocks share weights, effectively doubling the total number of layers. 

\section{Compatibility with Quantization}
\label{sec:appendix_quantization}
This section explores the compatibility of quantization with the proposed model architecture and layer sharing. We employ a straightforward per-token min-max quantization, quantizing both weight and activation to 8-bit using post-training quantization (PTQ). Experiments are conducted on both $\ours{}$ and $\ourss{}$ with model sizes of 125M and 350M, trained on 0.25T tokens. Results in Table~\ref{tab:appendix_ptq} demonstrate that W8A8 PTQ results in a modest accuracy drop of within 0.5\% and is compatible with the proposed layer-sharing method.
\begin{table*}[h]
\renewcommand\arraystretch{0.6}
\centering
\caption{Ablation study: 8-bit weight, 8-bit activation post-training quantization results on zero-shot common sense reasoning tasks. Quantized models achieve accuracy gap of within 0.5\% compared to the full-precision BF16 counterpart.}
\label{tab:appendix_ptq}
\setlength{\tabcolsep}{1mm}
{\resizebox{0.99\textwidth}{!}{
\begin{tabular}{lccccccccccccc}
\noalign{\vspace{0.1em}}\hline\noalign{\vspace{0.1em}}
\noalign{\vspace{0.1em}}\hline\noalign{\vspace{0.1em}}
\textbf{Model} & \textbf{Precision} & \textbf{ARC-e} & \textbf{ARC-c} & \textbf{BoolQ} & \textbf{PIQA} & \textbf{SIQA} & \textbf{HellaSwag} & \textbf{OBQA} & \textbf{WinoGrande} & \textbf{Avg.} & \textbf{Gap} \\ 
\noalign{\vspace{0.1em}}\hline\noalign{\vspace{0.1em}}
$\ours{}$-125M & BF16 & 45.5 & 27.7 & 58.3 & 64.6 & 41.9 & 36.4 & 35.4 & 50.4 & 45.0 & -- \\
$\ours{}$-125M  & W8A8 & 45.2 & 27.1 & 58.3 & 65.0 & 41.7 & 36.2 & 33.6 & 51.0 & 44.8 & 0.2  \\
\noalign{\vspace{0.1em}}\hdashline[0.8pt/1pt]\noalign{\vspace{0.2em}} 
$\ourss{}$-125M & BF16 & 44.4 & 27.0 & 61.5 & 65.1 & 43.0 & 37.6 & 37.8 & 52.0 & 46.1 & -- \\
$\ourss{}$-125M  & W8A8 & 44.0 & 27.5 & 60.9 & 64.6 & 43.1 & 37.7 & 37.7 & 51.0 & 45.8 & 0.3 \\
\noalign{\vspace{0.1em}}\hdashline[0.8pt/1pt]\noalign{\vspace{0.2em}} 
$\ours{}$-350M & BF16 & 51.4 & 31.3 & 61.0 & 68.1 & 43.6 & 47.2 & 41.6 & 55.4 & 49.9 & -- \\
$\ours{}$-350M  & W8A8 & 51.4 & 32.1 & 61.1 & 68.8 & 43.1 & 47.1 & 40.6 & 55.1 & 49.9 & 0.0 \\
\noalign{\vspace{0.1em}}\hdashline[0.8pt/1pt]\noalign{\vspace{0.2em}} 
$\ourss{}$-350M & BF16 & 51.9 & 35.2 & 59.6 & 68.9 & 43.4 & 47.2 & 43.3 & 58.4 & 51.0 & -- \\
$\ourss{}$-350M & W8A8 & 51.3 & 33.8 & 59.5 & 69.1 & 43.7 & 47.2 & 43.0 & 57.0 & 50.6 & 0.4 \\
\noalign{\vspace{0.1em}}\hline\noalign{\vspace{0.1em}}
\noalign{\vspace{0.1em}}\hline\noalign{\vspace{0.1em}}
\end{tabular}}}
\end{table*}

\section{Knowledge Distillation}
\label{sec:appendix_kd}
The results of integrating knowledge distillation (KD)~\cite{hinton2015distilling} into small model pre-training are presented in Table~\ref{tab:appendix_kd}. LLaMA-v2 7B models serve as the teacher, and the KD loss is computed using cross-entropy between the logits from the large pre-trained teacher model (\textit{i.e.}, LLaMA-v2 7B) and the small student network (\textit{i.e.}, 125M or 350M models): 
\begin{align}
    \mathcal{L}_{CE} = -\frac{1}{n}\sum_c\sum^n_{i=1} p_c^{\mathcal{T}}(X_i)\log(p_c^{\mathcal{S}}(X_i)),
\end{align}
Here, $i$ denotes the $i^{th}$ sample in the current batch with $n$ total samples in the batch, and $c$ represents the number of classes, which, in our case, equals the size of the vocabulary. $\mathcal{T}$ and $\mathcal{S}$ are the teacher and student networks, respectively.

The results in Table~\ref{tab:appendix_kd} indicate that adding KD loss is comparable to or even lower than solely using the next token as labels. However, it's noteworthy that the training time using KD is $2.6-3.2 \times$ slower than training from scratch using labels. All models are trained on 32 A100 80G GPUs with a batch size of 32 for 120k iterations. Consequently, we opt to use labels in our experiments.
\begin{table*}[h]
\renewcommand\arraystretch{0.6}
\centering
\caption{Ablation study on employing LLaMA-v2 7B teacher's output as soft labels for knowledge distillation (KD). Results indicate a slight degradation in performance when incorporating KD loss compared to only using hard labels. }
\label{tab:appendix_kd}
\setlength{\tabcolsep}{1mm}
{\resizebox{0.99\textwidth}{!}{
\begin{tabular}{lccccccccccccc}
\noalign{\vspace{0.1em}}\hline\noalign{\vspace{0.1em}}
\noalign{\vspace{0.1em}}\hline\noalign{\vspace{0.1em}}
\textbf{Model} & \textbf{Training Loss} & \textbf{Training Time} & \textbf{ARC-e} & \textbf{ARC-c} & \textbf{BoolQ} & \textbf{PIQA} & \textbf{SIQA} & \textbf{HellaSwag} & \textbf{OBQA} & \textbf{WinoGrande} & \textbf{Avg.} \\ 
\noalign{\vspace{0.1em}}\hline\noalign{\vspace{0.1em}}
125M model & Label & 29h & 43.1 & 28.9 & 58.1 & 62.3 & 42.3 & 34.6 & 31.5 & 50.1 & 43.9 \\ 
125M model & Label + KD & 93h & 41.8 & 28.5 & 58.5 & 61.6 & 41.1 & 34.5 & 32.7 & 51.6 & 43.8 \\ 
\noalign{\vspace{0.1em}}\hdashline[0.8pt/1pt]\noalign{\vspace{0.2em}} 
350M model & Label & 42h & 50.2 & 31.8 & 56.9 & 67.7 & 44.3 & 45.8 & 40.8 & 55.5 & 49.1 \\ 
350M model & Label + KD & 109h & 48.7 & 31.8 & 60.7 & 67.4 & 43.2 & 45.9 & 38.9 & 53.7 & 48.8 \\ 
\noalign{\vspace{0.1em}}\hline\noalign{\vspace{0.1em}}
\noalign{\vspace{0.1em}}\hline\noalign{\vspace{0.1em}}
\end{tabular}}}
\end{table*}

\section{Datasets and Benchmarks}
\label{sec:dataset}
$\ours{}$ is assessed on zero-shot common sense reasoning (BoolQ, PIQA, SIQA, HellaSwag, Winogrande, ARC, OBQA), question answering (TriviaQA), and reading comprehension (RACE) tasks. Additionally, we evaluate our chat models on MT-Bench and AlpacaEval benchmarks. We also generated an API calling dataset to fine-tune and evaluate models for this particular task.

\subsection{Zero-shot Common Sense Reasoning tasks}
\textbf{BoolQ}~\cite{clark2019boolq} is a reading comprehension dataset focused on naturally occurring yes/no questions. Each instance includes a question (Q), an excerpt from a passage (P), and an answer (A), with an added explanation for enhanced clarity.

\textbf{PIQA}~\cite{bisk2020piqa}, abbreviated for Physical Interaction: Question Answering, serves as a benchmark for evaluating and studying the capacity of natural language models in comprehending physical commonsense understanding.

\textbf{SIQA}~\cite{sap2019siqa}, abbreviated for Social Interaction Question Answering, is designed to measure the social and emotional intelligence of computational models through multiple-choice question answering.

\textbf{HellaSwag}~\cite{zellers2019hellaswag} serves as a benchmark for physically situated commonsense natural language inference. It comprises four-way multiple-choice problems that are considered trivial for humans (>95\% accuracy) but pose a challenge for language models.

\textbf{WinoGrande}~\cite{sakaguchi2021winogrande} is a benchmark for commonsense reasoning. It consists of a set of 273 expert-crafted pronoun resolution problems, deliberately designed to be unsolvable for statistical models relying on selectional preferences or word associations.

\textbf{ARC}~\cite{clark2018arc}, the AI2 Reasoning Challenge, is a compilation of 7787 natural science questions. It is divided into a Challenge Set and an Easy Set, with the Challenge Set exclusively comprising questions that were answered incorrectly by both a retrieval-based algorithm and a word co-occurrence algorithm.

\textbf{OBQA}~\cite{mihaylov2018obqa} is a dataset consisting of approximately 6000 questions designed for open-book question answering. The task involves integrating a corpus of provided science facts (open book) with external broad common knowledge. Providing correct answers necessitates leveraging simple common knowledge beyond the core facts provided.

\subsection{Question Answering Tasks}
\textbf{TriviaQA}~\cite{joshi2017triviaqa} is a closed-book question answering benchmark. It encompasses over 650,000 question-answer evidence triples derived by combining 95,000 question-answer pairs authored by Trivia enthusiasts. Each question is supported by an average of six evidence documents.

\subsection{Reading Comprehension Tasks}
\textbf{RACE}~\cite{lai2017race} is a large-scale reading comprehension dataset with more than 28,000 passages and nearly 100,000 questions. The dataset is collected from English examinations in China, which are designed for middle school and high school students.

\subsection{Chat Benchmarks}
\textbf{AlpacaEval}~\cite{alpaca_eval} is a single-turn benchmark featuring 805 questions across various topics. In the evaluation of this paper, models generate responses and are scored by GPT-4, with the ultimate metric being the pairwise win rate against a baseline model (text-davinci-001).

\textbf{MT-Bench}~\cite{zheng2023mtbench} is a multi-turn benchmark comprising 160 questions spanning eight diverse knowledge domains. In this evaluation, models are tasked with answering an initial question and subsequently providing a second response to a predefined follow-up question. Each model response is rated by GPT-4 on a scale from 1 to 10, and the final score is determined by the mean over the two turns.

\subsection{API Calling Dataset}
The API calling dataset is generated by instructing a language model to simulate a conversation between a human and an agent. This process instructs the language model to explicitly mention tasks that should be performed using designated APIs throughout the simulated conversation.
Examples of a conversation turn in the generated dataset are shown in the following boxes.

\begin{framed}
\centering
\begin{minipage}{.9\textwidth}
\{"role": "USER", "message": "Can you help me send a SMS message: We are looking forward to seeing you"\},\\
\{"role": "PARSER", "message": "style\_transfer(text = "We are looking forward to seeing you", style = ["sms"], contact = None)"\},\\
\{"role": "AGENT", "message": "Sure! Here's a possible SMS message: Hey [name], we are looking forward to seeing you! Wishing you all well."\},
\end{minipage}
\vspace{.5em}
\end{framed}

\begin{framed}
\centering
\begin{minipage}{.9\textwidth}
\{"role": "USER", "message": "How did the S\&P 500 perform last month?"\},\\
\{"role": "PARSER", "message": "get\_stock\_information(company\_name = "S\&P 500", ticker\_symbol = "GSPC", date\_time = "last month", financial\_metrics = "price")"\},\\
\{"role": "AGENT", "message": "The S\&P 500 closed at \$2,977.94 last month, with a monthly return of 2.5\%."\},
\end{minipage}
\vspace{.5em}
\end{framed}

\begin{framed}
\centering
\begin{minipage}{.9\textwidth}
\{"role": "USER", "message": "Is the San Diego Zoo open during the winter months?"\}, \\
\{"role": "PARSER", "message": "get\_local\_information(point\_of\_interest = "San Diego Zoo", attributes = "hours of operation")"\}, \\
\{"role": "AGENT", "message": "The San Diego Zoo is open year-round, but the hours of operation vary depending on the time of year. During the winter months, the zoo is open from 9:00 AM to 5:00 PM."\},
\end{minipage}
\vspace{.5em}
\end{framed}
\begin{framed}
\centering
\begin{minipage}{.9\textwidth}
\{"role": "USER", "message": "What is the most recent news concerning aliens?"\}, \\
\{"role": "PARSER", "message": "news(query = "latest alien news")"\}, \\
\{"role": "AGENT", "message": "According to recent news articles, NASA has announced the discovery of a new planet outside of our solar system that has the potential to support alien life. The planet, named Kepler-186f, is located in the constellation Cygnus and is the first known Earth-sized planet to be found in the habitable zone of a star. While this news does not necessarily prove the existence of aliens, it does provide hope for the possibility of finding life beyond Earth. Additionally, a recent study published in the journal Nature has suggested that the universe may be teeming with alien civilizations, with estimates suggesting that there could be tens of billions of intelligent civilizations in the observable universe. However, it's important to note that these findings are still being researched and debated in the scientific community."\},
\end{minipage}
\end{framed}

\begin{framed}
\centering
\begin{minipage}{.9\textwidth}
\{"role": "USER", "message": "Who won the last Super Bowl?"\}, \\
\{"role": "PARSER", "message": "get\_sports\_information(match = "", sport\_type = "NFL", metric = "winner")"\}, \\
\{"role": "AGENT", "message": "The last Super Bowl was won by the Kansas City Chiefs."\},
\end{minipage}
\vspace{.5em}
\end{framed}
\newpage

\section{Approximate calculation of envisioned future GPU consumption}
\#GPUs = $7.88 \!\!\times\!\! 10^9$ (population size)  $\times 5\%$ (individual time\footnote{According to the statistics that YouTube users spend an average of 23 hours per month (\href{https://www.mixbloom.com/resources/average-time-spent-on-social-media-2022}{source}) and Instagram users dedicating 12 hours monthly to the app (\href{https://blog.hootsuite.com/instagram-statistics/}{source}), we have reason to believe the interaction with LLM can easily surpass the cumulative engagement of these platforms in the future.})  $\times \ 220 \!\!\times\!\! 10^9 $ (approximating FLOPs/token for 220B GPT-4 model with one expert activated\footnote{https://the-decoder.com/gpt-4-has-a-trillion-parameters/}) $\times 50 $ (token/s) $\times 24$ (h) $\times 3600$ (s) / ($60 \!\!\times\!\! 10^{12} $ (FLOPs/s, H100 computation capacity ) $\times 24$ (h) $\times 3600$ (s)) $\approx$ $1\times10^8$

This assumes short sequence lengths (\textit{i.e.} hundreds of tokens, not tens of thousands of tokens), where the compute is dominated by linear layers.
For very long sequences, the self-attention computation would dominate, and this would necessitate a more complicated formula.

\section{Chat Generations from \ours{}}
\label{sec:chat_generation}
We show several examples of chat output generated with $\ours{}$ while responding to various questions.

\centering
\begin{minipage}{.95\textwidth}
\begin{framed}
\textbf{Input:}\\Are there any particular physical benefits to mindful walking, such as improved posture or increased physical fitness?
\end{framed}
\end{minipage}

\centering
\begin{minipage}{.95\textwidth}
\begin{framed}
\textbf{Output} ($\ours$-125M):\\Yes, there are several physical benefits to mindful walking. Here are a few:

1. Improved Posture: Mindful walking can help improve your posture by promoting good alignment and reducing stress on the joints. This can lead to better posture and reduced back pain.

2. Increased Flexibility: Mindful walking can also help increase flexibility in your body, which can reduce the risk of injury and improve overall health.

3. Better Sleep: Mindful walking can help you fall asleep more easily and stay asleep longer, which can improve your sleep quality.

4. Reduced Stress: Mindful walking can help reduce stress levels by promoting relaxation and reducing tension in the body.

5. Enhanced Mental Health: Mindful walking has been shown to have positive effects on mental health, including reducing symptoms of depression and anxiety.

Overall, mindful walking is an excellent way to improve your physical health and well-being. It's a simple and effective way to get moving and enjoy the benefits it offers. If you're interested in trying mindful walking, I would recommend starting with shorter walks and gradually increasing the duration over time. You may also want to consider incorporating mindfulness into your daily routine to further enhance your experience.

In conclusion, mindful walking is a great way to improve your physical health and well-being. With its numerous physical benefits, it's easy to see why it's becoming increasingly popular among people looking for a fun and sustainable way to get moving. So go ahead and give it a try!
\end{framed}
\end{minipage}

\begin{minipage}{.95\textwidth}
\begin{framed}
\textbf{Output} ($\ours$-350M):\\
Yes, there are several physical benefits of mindful walking. Firstly, it can improve your posture by reducing the amount of tension in your muscles and joints. Secondly, it can increase your physical fitness by improving your cardiovascular health, strengthening your core muscles, and increasing your flexibility. Additionally, mindful walking can help you to stay more focused on the present moment, which can lead to better sleep quality and reduced stress levels. Finally, it can also help you to develop a greater sense of awareness and appreciation for the world around you, which can enhance your overall well-being.

Overall, mindful walking is a great way to improve your physical health and well-being, while also promoting relaxation and mindfulness. Whether you're looking to reduce stress, improve your posture, or simply enjoy a peaceful walk, mindful walking can be a rewarding and fulfilling experience.

If you're interested in trying out mindful walking, there are many resources available that can guide you through the process. Some popular options include guided meditations, apps that offer mindful walking exercises, and classes that teach you how to practice mindful walking. With a little bit of effort and dedication, anyone can start incorporating mindful walking into their daily routine and reap the many benefits it offers.

In conclusion, mindful walking is a simple yet powerful tool that can help you to improve your physical health, reduce stress, and cultivate a deeper sense of awareness and appreciation for the world around you. So why not give it a try and see what kind of benefits you can achieve? Your body and mind will thank you!
\end{framed}
\end{minipage}

\begin{minipage}{.95\textwidth}
\begin{framed}
\textbf{Output} (llama-v2-7b-chat):\\
Mindful walking can have a range of physical benefits, including:

1. Improved posture: By focusing on the sensation of each step and the movement of the body, mindful walking can help improve posture and reduce slouching or hunching.

2. Increased physical fitness: Mindful walking can help improve cardiovascular fitness, increase muscle strength and endurance, and boost overall physical fitness.

3. Better balance and coordination: By paying attention to the body's position in space and the movement of the legs and feet, mindful walking can help improve balance and coordination.

4. Reduced stress and tension: Mindful walking can help reduce stress and tension in the body, which can lead to improved physical relaxation and well-being.

5. Improved flexibility: By focusing on the movement of the body and the sensation of each step, mindful walking can help improve flexibility and range of motion.

6. Boosted immune system: Studies have shown that mindfulness practices, including mindful walking, can help boost the immune system and reduce inflammation.

7. Improved sleep: Mindful walking can help improve sleep quality and duration by reducing stress and promoting relaxation.
8. Reduced pain: Mindful walking can help reduce pain by focusing on the sensation of each step and the movement of the body, rather than the pain itself.

9. Improved mood: Mindful walking can help improve mood and reduce symptoms of depression and anxiety by promoting relaxation and reducing stress.

It's important to note that the physical benefits of mindful walking may vary depending on the individual and the frequency and duration of the practice. However, with regular practice, mindful walking can have a range of positive effects on physical health and well-being.
\end{framed}
\end{minipage}
\vspace{2em}

\centering
\begin{minipage}{.95\textwidth}
\begin{framed}
\textbf{Input:}\\
How do I wrap a present neatly?
\end{framed}
\end{minipage}

\centering
\begin{minipage}{.95\textwidth}
\begin{framed}
\textbf{Output} ($\ours$-125M):\\
1. Choose the right size gift: The size of the present you want to give will depend on your budget and personal preference. If you are looking for a small gift, consider buying a small box or card with a single-use plastic bag.

2. Prepare the present: Make sure all the items in the present are clean and tidy. You can use a cleaning kit or a tissue paper to remove any dirt or debris from the present.

3. Arrange the presents: Place the present in a place where it is easy to see and touch. Use a pencil to draw a line around the present and arrange it in a way that makes sense to you.

4. Add decorations: Decorate the present with stickers, ribbons, or other decorative items. This will add some personality to the present and make it more visually appealing.

5. Enjoy the present: Take time to enjoy the present and let yourself feel the warmth and comfort of the present. Smile at yourself and take a moment to appreciate the present.

Remember, the most important thing is to have fun and enjoy the present! Happy gift-giving!

If you're feeling adventurous, try making a surprise gift for someone special by wrapping the present in a cute and colorful gift basket. Or, if you're feeling adventurous, try making a gift for a friend who needs a little extra help with their homework. Whatever you choose, remember to be creative and have fun with your present!

With these tips, you'll be able to create a beautiful and thoughtful present that will last for years to come. Happy gift-giving!
\end{framed}
\end{minipage}

\centering
\begin{minipage}{.95\textwidth}
\begin{framed}
\textbf{Output} ($\ours$-350M):\\
1. Start by wrapping the present in a decorative gift wrap or ribbon.

2. Use a small amount of glue or tape to secure the gift wrap or ribbon to the surface of the present.

3. Place the gift wrap or ribbon around the present, making sure it is centered and evenly distributed.

4. Use a small amount of glue or tape to secure the gift wrap or ribbon to the surface of the present.

5. Once the gift wrap or ribbon is securely attached, you can add any additional decorations or embellishments to the present.

6. If you want to add a personal touch, you can write a message or a special message on the gift wrap or ribbon.

7. Finally, enjoy your beautifully wrapped present!

Remember, wrapping a gift is a fun and creative way to show your loved one how much you care. Have fun and enjoy the process!
\end{framed}
\end{minipage}

\centering
\begin{minipage}{.95\textwidth}
\begin{framed}
\textbf{Output} (llama-v2-7b-chat):\\
Wrapping a present neatly can be a bit of a challenge, but with a few simple tips and tricks, you can create a beautifully wrapped gift that your recipient will love. Here are some tips to help you wrap a present neatly:

1. Use the right wrapping paper: Choose a wrapping paper that complements the gift and is the right size for the present. You can also use a patterned paper or a solid color, depending on your preference.

2. Measure the gift: Before you start wrapping, measure the gift to make sure you have enough wrapping paper. You can also use a piece of scrap paper to measure the gift and mark the edges with a pencil.

3. Use a flat surface: Make sure you have a flat surface to work on, such as a table or countertop. This will help you keep the wrapping paper flat and prevent it from wrinkling.

4. Start with the edges: Begin by folding the edges of the wrapping paper over the gift, making sure they are even and smooth. You can use tape to secure the edges if needed.

5. Roll the gift: Once you have the edges in place, roll the gift into the center of the wrapping paper. Make sure the gift is securely in place before moving on to the next step.

6. Tuck in the edges: As you wrap the gift, tuck the edges of the wrapping paper under the gift, smoothing them out as you go. Make sure the edges are even and smooth, and use tape to secure them if needed.

7. Add a ribbon or bow: Once you have the gift wrapped, add a ribbon or bow to the top. This will not only make the gift look more festive, but it will also help to keep the wrapping paper in place.

8. Practice makes perfect: Wrapping a present neatly takes a bit of practice, so don't be discouraged if it doesn't come out perfectly the first time. Keep practicing, and you'll get the hang of it in no time!

By following these tips, you can create a beautifully wrapped present that your recipient will love. Happy wrapping!
\end{framed}
\end{minipage}
\vspace{2em}

\centering
\begin{minipage}{.95\textwidth}
\begin{framed}
\textbf{Input:}\\
What are the best exercises for beginners?
\end{framed}
\end{minipage}

\centering
\begin{minipage}{.95\textwidth}
\begin{framed}
\textbf{Output} ($\ours$-125M):\\
1. Push-ups: This is a classic exercise that can be done anywhere, anytime. It's easy to do and requires no equipment.

2. Dumbbell Rows: These rows are great for building strength in your legs and core.

3. Squats: Squats are a great way to build strength in your legs and core.

4. Lunges: Lunges are a great way to improve balance and stability.

5. Plank: A plank is a great way to build strength in your back and core.

6. Jumping Jacks: Jumping jacks are a fun and effective way to build strength in your legs and core.

7. Burpees: Burpees are a great way to build strength in your legs and core.

8. Mountain Climbers: Mountain climbers are a great way to build strength in your legs and core.

9. Deadlifts: Deadlifts are a great way to build strength in your legs and core.

10. Tricep Dips: Tricep dips are a great way to build strength in your triceps.

Remember, it's important to start with simple exercises and gradually increase the intensity as you get stronger. Don't push yourself too hard or you may find it difficult to maintain proper form. With consistent practice, you will see improvements in your strength and overall fitness.

If you have any questions about exercises or healthy living, don't hesitate to talk to a doctor or physical therapist. They can help you develop a personalized plan that works best for you.

In conclusion, there are many exercises that can be done at home, but it's important to start slowly and gradually increase the intensity over time. Remember to listen to your body and take breaks when needed. And most importantly, enjoy the process!
\end{framed}
\end{minipage}

\centering
\begin{minipage}{.95\textwidth}
\begin{framed}
\textbf{Output} ($\ours$-350M):\\
Beginners should start with exercises that are easy to do and don't require a lot of strength or endurance. Here are some exercises that are great for beginners:

1. Push-ups: This is a great exercise for beginners because it's a simple exercise that can be done anywhere. Start with a push-up on your knees and then lower yourself down to the ground.

2. Squats: Squats are a great exercise for building leg strength and endurance. Start with a squat on your knees and then lower yourself down to the ground.

3. Lunges: Lunges are a great exercise for building leg strength and endurance. Start with a lunge on your front foot and then lower yourself down to the ground.

4. Plank: Planks are a great exercise for building core strength and stability. Start with a plank on your forearms and then lower yourself down to the ground.

5. Bicycle crunches: Bicycle crunches are a great exercise for building upper body strength and stability. Start with a bicycle crunch on your upper back and then lower yourself down to the ground.

6. Plank with leg lifts: This exercise is great for building leg strength and stability. Start with a plank on your forearms and then lift your legs off the ground and hold for a few seconds before lowering them back down.

7. Jumping jacks: Jumping jacks are a great exercise for building cardiovascular endurance and strength. Start with a jumping jack on your feet and then jump up and spread your legs.

Remember to start with a weight that is comfortable for you and gradually increase the weight as you get stronger. It's also important to warm up before exercising and to stretch after your workout to prevent injury.

In conclusion, beginners should start with exercises that are easy to do and don't require a lot of strength or endurance. These exercises are great for building leg strength and endurance, and can be done anywhere. Remember to start with a weight that is comfortable for you and gradually increase the weight as you get stronger. Warming up and stretching after your workout are also important to prevent injury.

Following these exercises can help you build strength, endurance, and improve your overall fitness level. Remember to always consult with a healthcare professional before starting any new exercise program.
\end{framed}
\end{minipage}

\centering
\begin{minipage}{.95\textwidth}
\begin{framed}
\textbf{Output} (llama-v2-7b-chat):\\
Exercise is an essential part of a healthy lifestyle, but for beginners, it can be intimidating and overwhelming. Here are some of the best exercises for beginners:

1. Bodyweight exercises: These exercises are great for beginners because they don't require any equipment. Examples include push-ups, squats, lunges, and planks.

2. Yoga: Yoga is a low-impact exercise that can help improve flexibility, balance, and strength. It's also great for reducing stress and improving overall well-being.

3. Brisk walking: Walking is a simple and accessible form of exercise that can be done anywhere. It's a great way to get your heart rate up and improve cardiovascular health.

4. Swimming: Swimming is a low-impact exercise that can help improve cardiovascular health and strengthen muscles. It's also easy on the joints, making it a great option for beginners.

5. Cycling: Cycling is another low-impact exercise that can be done indoors or outdoors. It's a great way to improve cardiovascular health and build leg strength.

6. Resistance band exercises: Resistance bands are lightweight and portable, making them easy to use at home or on the go. They can help improve strength and flexibility in the upper body, lower body, and core.

7. Pilates: Pilates is a form of exercise that focuses on core strength, flexibility, and body control. It's a great option for beginners because it's low-impact and can help improve posture and balance.

8. Dance-based workouts: Dance-based workouts, such as Zumba or Hip Hop Abs, are fun and energetic ways to get in shape. They often involve aerobic exercise and strength training, making them a great option for beginners.

9. Jumping rope: Jumping rope is a high-intensity exercise that can help improve cardiovascular health and burn calories. It's also easy to do at home with a simple jump rope.

10. Kettlebell exercises: Kettlebells are small weights that can be used to perform a variety of exercises, including swings, squats, and presses. They're a great way to improve strength and cardiovascular health.

Remember to always consult with a doctor before starting any new exercise program, especially if you're a beginner. It's also important to start slowly and gradually increase the intensity and duration of your workouts as you become more comfortable.
\end{framed}
\end{minipage}
\vspace{2em}

\end{document}